\documentclass[conference]{IEEEtran}
\IEEEoverridecommandlockouts
\usepackage{natbib}
\usepackage{amsmath,amssymb,amsfonts}
\usepackage{algorithmic}
\usepackage{graphicx}
\usepackage{textcomp}
\usepackage{xcolor}
\usepackage[]{mdframed}
\usepackage{float}    
\usepackage{listings}
\usepackage{caption}
\usepackage{subcaption}
\usepackage{multicol}
\usepackage{multirow}  
\usepackage{hyperref}  
\usepackage{booktabs}  

\usepackage{xurl}

\def\BibTeX{{\rm B\kern-.05em{\sc i\kern-.025em b}\kern-.08em
    T\kern-.1667em\lower.7ex\hbox{E}\kern-.125emX}}

\usepackage{xcolor}

\definecolor{codegreen}{rgb}{0,0.6,0}
\definecolor{codegray}{rgb}{0.5,0.5,0.5}
\definecolor{codepurple}{rgb}{0.58,0,0.82}
\definecolor{backcolour}{rgb}{0.95,0.95,0.92}

\lstdefinestyle{mystyle}{
    backgroundcolor=\color{backcolour},   
    commentstyle=\color{codegreen},
    keywordstyle=\color{magenta},
    numberstyle=\tiny\color{codegray},
    stringstyle=\color{codepurple},
    basicstyle=\ttfamily\footnotesize,
    breakatwhitespace=false,         
    breaklines=true,                 
    captionpos=b,                    
    keepspaces=true,                 
    numbers=left,                    
    numbersep=5pt,                  
    showspaces=false,                
    showstringspaces=false,
    showtabs=false,                  
    tabsize=2
}

\newcommand \codebert {CodeBERT}
\newcommand \codegraph {CodeGraph}
\newcommand \codetwovec {Code2Vec}

\newcommand \bcb {BCB}
\newcommand \poolc {PoolC}
\newcommand \mixone {Mix\_1}

\begin{document}

\title{Source Code is a Graph, Not a Sequence: A Cross-Lingual Perspective on Code Clone Detection \\
{\footnotesize \textsuperscript{*}A Challenge and Opportunity for Source Code Research: Through the Lens of Code Clone Detection using Python and Java.}
\thanks{Source code can be found here \url{https://github.com/Ataago-AI/clone-detection}}
}

\author{
    \IEEEauthorblockN{Mohammed Ataaur Rahaman}
    \IEEEauthorblockA{
        \textit{School of Electronic Engineering and Computer Science} \\
        \textit{Queen Mary University}\\
            London, UK \\
            m.a.rahaman@se22.qmul.ac.uk
        }
\and
    \IEEEauthorblockN{Julia Ive}
    \IEEEauthorblockA{
        \textit{School of Electronic Engineering and Computer Science} \\
        \textit{Queen Mary University}\\
            London, UK \\
            j.ive@qmul.ac.uk
        }
        
}

\maketitle

\begin{abstract}

Source code clone detection is the task of finding code fragments that have the same or similar functionality, but may differ in syntax or structure. This task is important for software maintenance, reuse, and quality assurance \citep{SCA_comparision_n_eval_of_CD}. However, code clone detection is challenging, as source code can be written in different languages, domains, and styles. In this paper, we argue that source code is inherently a graph, not a sequence, and that graph-based methods are more suitable for code clone detection than sequence-based methods. We compare the performance of two state-of-the-art models: \codebert{} \citep{CodeBERT}, a sequence-based model, and \codegraph{} \citep{CodeGraph4CCDetector}, a graph-based model, on two benchmark data-sets: \bcb{} \citep{BCB_original} and \poolc{} \citep{PoolC}. We show that \codegraph{} outperforms \codebert{} on both data-sets, especially on cross-lingual code clones. To the best of our knowledge, this is the first work to demonstrate the superiority of graph-based methods over sequence-based methods on cross-lingual code clone detection.
\end{abstract}


\section{Introduction}
Source code is a formal language that describes the logic and behavior of a software system \citep{computational_thinking}. Source code can be written in different programming languages, domains, and styles, but it always has a common characteristic, i.e. it is structured as a graph, not a sequence. A graph is a mathematical object that consists of nodes and edges, where nodes represent entities and edges represent relations \citep{intro_to_graph_theory}. A graph can capture both the syntactic and semantic information of the source code, such as the tokens, statements, control flow, data flow, and program dependence to name a few. To realise that source code is a graph, not a sequence, we take a simple code task of identifying if a pair of source code given is a clone or not. 

Code clone detection is the task of finding code fragments that are identical or similar in functionality, but may differ in syntax or structure. Code clone detection is important for software engineering, as it can help to improve code quality, reduce maintenance costs, and prevent bugs and vulnerabilities \citep{SCA_comparision_n_eval_of_CD}. However, code clone detection is also challenging, as it requires a deep understanding of the semantics and logic of the code, which may vary across different programming languages, domains, and styles.

Existing methods for code clone detection can be broadly classified into two categories: sequence-based and graph-based. Sequence-based methods rely on textual similarity of the code, such as token sequences. Graph-based methods rely on structural similarity of the code, such as Abstract Syntax Tree (ASTs), or control flow graphs (CFGs) or Code Property Graphs (CPGs). Sequence-based methods are fast and scalable, but they may fail to detect clones that have different syntax or structure. Graph-based methods are more accurate and robust, but they may be slow and complex, especially for large-scale or cross-language code clone detection.

A python source code clone pair is presented in Listing [\ref{motivational_expample_1}, \ref{motivational_expample_2}]. The two code snippets have the same semantic behavior: they print ‘A’ or ‘a’ depending on the case of the input. However, they differ in their syntactic forms. Further such examples can be viewed in Appendix \ref{appendix:code_rep}.

\begin{multicols}{2}[]
\centering
\begin{lstlisting}[language=Python, caption=Python code 1, linewidth=.99\columnwidth, label=motivational_expample_1]
S = input()

if S.isupper():
  print("A")
else:
  print("a")\end{lstlisting}
\begin{lstlisting}[language=Python, caption=Python code 2, linewidth=.99\columnwidth, label=motivational_expample_2]
alp=input()

if alp==alp.upper():
  print("A")
elif alp==alp.lower():
  print("a")\end{lstlisting}
\end{multicols}

In this paper, we argue that source code is naturally a graph, not a sequence, and that graph-based methods are more suitable for code clone detection than sequence-based methods. We compare the performance of sequence-based and graph-based methods for code clone detection on two benchmark data-sets: \bcb{} \citep{BCB_original} and \poolc{} \citep{PoolC}. \bcb{} is a data-set of Java code snippets where as \poolc{} is a data-set of Python code snippets. We use \codebert{} \citep{CodeBERT} as a representative sequence-based modelling approach, and \codegraph{} \citep{CodeGraph4CCDetector} as a representative graph-based modeling approach. \codebert{} is a bimodal pre-trained model for programming language (PL) and natural language (NL) that learns general-purpose representations that support downstream NL-PL applications. \codegraph{} is a graph-based model for semantic code clone detection based on a Siamese graph-matching network that uses attention mechanisms to learn code semantics from DFGs and CPGs.

We conduct various experiments to evaluate the accuracy, recall, precision, and F1-score of \codebert{} and \codegraph{} on three experimental setups: (i) in-domain static source code analysis , (ii) cross-lingual generalization and semantic extraction, and (iii) zero-shot source code clone classification. We show that \codegraph{} outperforms \codebert{} on all experimental setups in terms of all metrics. The main contributions of this paper are as follows:

\begin{itemize}
    \item To best of our knowledge, we are the first one to demonstrate the superiority of graph-based methods over sequence-based methods for multilingual static source code analysis tasks, such as clone detection, by exploiting the natural graph structure of source code across programming languages.
    \item We provide novel insights on the generalization and cross-domain understanding of graph-based models, compared to sequence-based models, for source code analysis, as they leverage both the syntactic and semantic features of source code in various cross-domain settings.
    \item We show how mixing cross-lingual data-sets can improve the overall performance of the graph-based model by 4.5\%, as it can learn from the commonalities and differences of different programming languages.
    \item We focus on the under-explored clone detection python data-set POOLC, along with the benchmark Java data-set BCB, and draw parallel comparisons on both of the data-sets.
    \item We provide highly efficient and scalable code for standard CPG representation generation using Tree-sitter \citep{tree_sitter} and Micrsoft's DFG \citep{GraphCodeBERT}, along with the re-implemented code for the sequence and graph-based models.
    
\end{itemize}

The rest of this paper is organized as follows: Section \ref{sec:related_work} reviews the related work on source code clone detection. Section \ref{sec:methods} introduces to the various code representations, like Abstract syntax tree, Data flow graph etc and the background knowledge on graph neural networks and natural language processing. Section \ref{sec:experimental_design} describes the details of how the experiments are carried out, and lays down the research questions. Section \ref{sec:results} presents the results and its discussion. Section \ref{sec:limitations_n_future_research} discusses the limitations and future research directions. Section \ref{sec:conclusion} concludes the paper.

\begin{figure*}[!ht]
    \centering
    \centerline{\includegraphics[width=\textwidth]{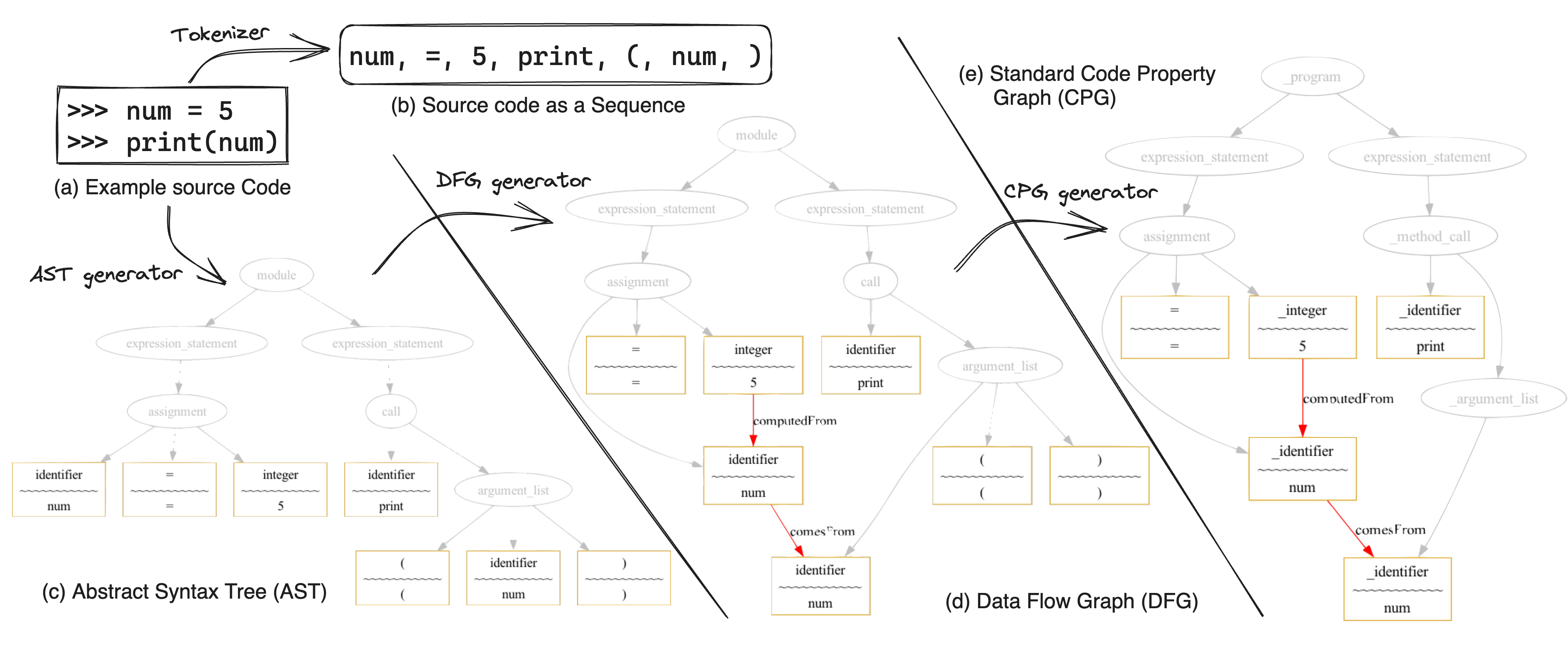}}
    \caption{This figure illustrates how the source code can be transformed into a sequence and a graph. (a) A sample Python program that prints hello, name. (b) The code tokens using CodeBERT’s BPE tokenizer. (c), (d), and (e) are the graph representations of the code as Abstract Syntax Tree, Data Flow Graph, and Standard Code Property Graph, respectively. This shows how Standard CPG (e) is the most concise and standardized graph representation across languages, compared to raw, AST, or DFG.}
    \label{fig:code_representations_001}
\end{figure*}

\section{Related Work}\label{sec:related_work}

\subsection{What is static source code analysis?}
Static code analysis is a valuable technique for improving software quality and security without actually compiling the code. It can find errors that are hard to detect at run time, improve the quality and maintainability of the code, and reduce the cost and time of testing and debugging. This is basically a form of white-box testing. According to \citet{Boehm} the cost of fixing a defect increases exponentially as it moves from the coding phase to the testing phase to the maintenance phase. Therefore, having a static tools to analyse and fix a source code as soon as possible helps save lot of resources and efforts. 

\subsection{What are applications of static code analysis?}
There are various different applications for static code analysis. Security Vulnerability detection, is one of the major static code analysis, which can help developers identify and fix security vulnerabilities before they are exploited by the attackers as extensively stated by \citet{SCA_SVD}. Another important area of static code analysis is in helping developers find inefficient algorithms and improve the resource utilization, response time, and other throughput of the software. Clone Detection is another application that can help identify similar functional code fragments which may indicate code duplication, plagiarism or reuse. This can improve quality, maintainability and the security of the code by eliminating redundant and inconsistent or outdated code \citep{SCA_CD}.

\subsection{What is source code clone detection?}
Source code clone detection is the process of finding code fragments that have similar functionalities or structures, which could indicate that the code is a duplicate, plagiarised or reused, which may be done purposefully, negligently or accidentally by a developer as said by \citet{SCA_CD}. Clone detection is very harmful to the quality of the entire source code \citep{SCA_CD}. A broad categorization on various types of code clones is given by \citep{SCA_CD}. 
\begin{itemize}
    \item \textbf{Textual Similarity}
    \begin{itemize}
        \item \textbf{Type I}: Changes in White-spaces, comments, layouts.
        \item \textbf{Type II}: Renaming of variable names, or changes in types and identifiers. 
        \item \textbf{Type III}: Addition or removal of statements.
    \end{itemize}
    \item \textbf{Functional Similarity}
    \begin{itemize}
        \item \textbf{Type IV}: Complete change in syntax, but functionally same behavior.
    \end{itemize}
\end{itemize}

\subsection{Types of Clone detection approaches?}
According to \citet{SCA_CD} there are multiple techniques to detect a source code clones, like Text based, token based, tree based, Program dependency graph based, metrics based, or hybrid approaches. Here we deal and compare token based vs tree based clone detection approach. As we know that, to detection semantically same code clones, the model should not just rely on difference of syntax, but also understand the semantics of the structure. This becomes hard, if we pass the source code to the model as sequence rather than a Tree like Abstract Syntax Tree which naturally holds the syntatical information of a source code.

\subsection{Sequence based modeling}
There has been various sequence based modelling approaches used by source code clone detection like, CodeBERT \citep{CodeBERT}, UNIXCODER \citep{Unixcoder}, ContraBERT \citep{ContraBERT}. Here in sequence modeling the source code is tokenized as a piece of words (or source code). This tokenized pieces of words in a sequence is learnt by the model to understand a fragment of code. This helps the model learn the semantics, by taking the code in a sequential manner. 

We use \codebert{} \citep{CodeBERT} which is a bimodal pre-trained model for programming language (PL) and natural language (NL) that learns general-purpose representations that support downstream NL-PL applications such as natural language code search, code documentation generation, etc1. \codebert{} is developed with a Transformer-based neural architecture, and is trained with a hybrid objective function that incorporates the pre-training task of replaced token detection, which is to detect plausible alternatives sampled from generators 1, along side with Masked Language modelling. In this study, we use CodeBERT as a pre-trained model for our sequence model for source code clone detection.

\subsection{Graph based modeling}
On the other side, clone detection as a graph modelling approach, we have models like TBCCD \citep{TBCCD}, FA-AST \citep{FA_AST_BCB_filterd}, HOLMES \citep{HOLMES},  DG-IVHFS \citep{DG_IVHFS}, CodeGraph4CCDetector \citep{CodeGraph4CCDetector}. These types of graph models first construct a tree or a graph like, abstract syntax tree, Control flow graph etc from the source code. This helps to retain the structural information of the code, regardless of it being moved from its location or variables being replaced. This ideally should help the model concentrate more on the semantics, rather than the structural learning, as it is already baked into its structure. 

We use CodeGraph4CCDetector \citep{CodeGraph4CCDetector} as our graph-based model, from here on referred as \codegraph{}. This model is reported to have state of the art results on the \bcb{} \citep{BCB_original} data-set. This is a Siamese graph matching network which basically takes in two source code snippets and output a similarity score between them. The input for this is the Code Property Graph, which is essentially graph having various nodes and edges. This helps the network capture the source codes syntactical and semantical information. The node representation of this \codegraph{} uses attention mechanism on a node level to extract out a node representation, before combining it to graph level representation. The major advantage of a graph level over the sequence level is, this can handle code snippets of different lengths and structures, as long as the hardware memory can load it.

\begin{figure*}[!ht]
    \centering
    \centerline{\includegraphics[width=\textwidth]{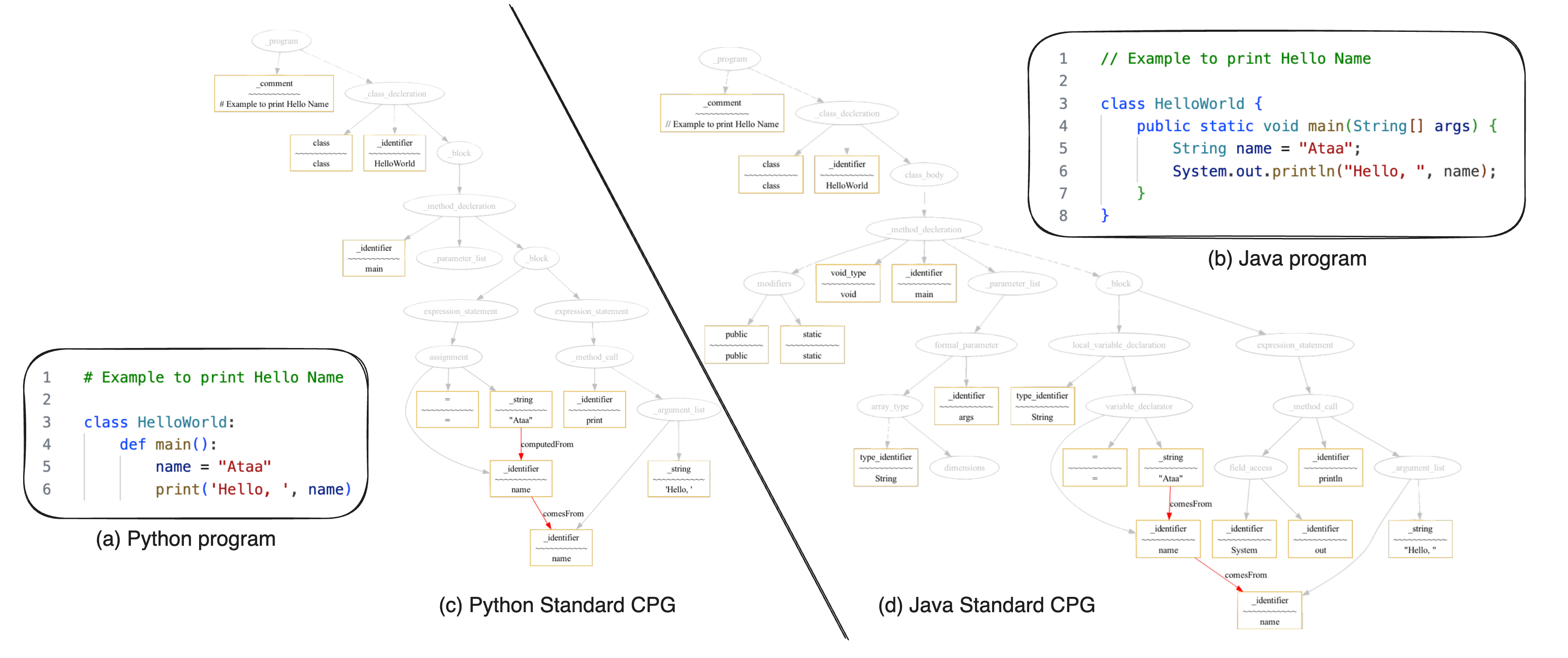}}
    \caption{(a) and (b) show the Java and Python programs that print hello, name, respectively. (c) and (d) show the corresponding standard CPGs, which are generated by applying lexical parsing, data flow generation, and graph standardization to the source code. The standard CPGs look identical for both languages, as they capture the common structure and logic of the programs.}
    \label{fig:code_representations_002}
\end{figure*}

\section{Methods}\label{sec:methods}

This section describes the methods and models that we employ for our experiments on the sequence and graph representations of source code. We organize this section into four parts. First, we present how we use byte pair tokenizers to represent source code as a sequence of tokens. Second, we illustrate how we use code property graphs (CPGs) to represent source code as a graph. Third, we introduce \codebert{} \citep{CodeBERT}, the sequence-based model that we employ for code clone detection. Fourth, we present \codegraph{} \citep{CodeGraph4CCDetector}, the graph-based model that we employ for code clone detection.

\subsection{Code Tokenization}
We apply the default Byte Pair Encoding, BPE tokenizer of CodeBERT \citep{CodeBERT} to represent the source code as a sequence of tokens. Figure \ref{fig:code_representations_001}b shows an example of how the BPE tokenizer splits the source code in Figure \ref{fig:code_representations_001}a into word and subword tokens.

\subsection{Code Representations}
We use a graph representation of source code that consists of nodes and edges connecting source code tokens. To generate this representation, we apply the following steps: First, we use a lexical parser, to produce the abstract syntax tree (AST) of the source code. Second, we extract the data flow information from the AST. Third, we merge the AST and DFG, to form one graph which we call it as Code Property Graph. This is further pruned and standardized across source code languages to make the standard CPG. 

\subsubsection{Abstract Syntax Tree (AST)}
We use Tree-sitter \citep{tree_sitter}, a lexical parser, to generate the abstract syntax tree (AST) of the source code for any language. We currently use it for Java and Python, but it supports 141 different languages. Figure \ref{fig:code_representations_001}c shows the AST generated from the sample source code in Figure \ref{fig:code_representations_001}a.

\subsubsection{Data Flow Graph (DFG)}
 We use Microsoft’s Data Flow Generator (DFG) \citep{GraphCodeBERT} to generate a data flow graph (DFG). This DFG generator takes the AST from the previous step and adds the data flow edges to it to form the DFG. Figure \ref{fig:code_representations_001}d shows the DFG graph for the same example source code in Figure \ref{fig:code_representations_001}a. We can see that there is a data flow edge between the integer literal ‘5’ and the variable ‘num’, as ‘5’ is assigned to ‘num’. There is also a data flow edge between the second occurrence of ‘num’ and the print statement, as ‘num’ is used as an argument.
 

\subsubsection{Standard Code Property Graph (CPG)}
We standardize the DFG to a code property graph (CPG), which is our final graph representation of source code. We perform two main operations to standardize the DFG across languages. First, we prune the graph from nodes that do not add value to the model’s understanding, such as opening and closing brackets that are implicitly understood when there is a method call. Second, we standardize the node type labels across languages so that the model can recognize them consistently across languages. For example, in Figure \ref{fig:code_representations_001}e we see that the root node 'module' and 'integer' node are standardized and replaced with ‘\_program’ and ‘\_integer’ as standard node types.


The major impact of a standard CPG can be seen in Figure \ref{fig:code_representations_002}, where two programs that print hello, name in Java and Python are written. The programs look different as raw code, but they have the same functionality and semantics. The standard CPGs look very similar in both cases, as shown in Figure \ref{fig:code_representations_002}c and \ref{fig:code_representations_002}d. We provide more examples of standard CPGs in Appendix \ref{appendix:code_rep}. This supports our claim that this type of code representation is more suitable than the sequence of code for identifying code clones.


\subsection{Sequence-based Model: \codebert{}}
In order to model a sequence model for source code clone detection, we use \codebert{} \citep{CodeBERT} as a pre-trained model. We fine-tune \codebert{} on the source code clone detection labelled data-set. The fine tuning task is a binary classification task where the source code pair is passed sequentially through the \codebert{} which acts as an encoder, and the 2 representation vectors coming out from this encoder, is concatenated and passed to a shallow 2 layer MLP classifier to give the final output, if the pair is a clone or a no clone. The major advantage of using this state of the art encoder \codebert{} is that it can help capture both the syntactic and semantic information of the PL code, by leveraging the large-scale pre-training data of multiple languages.



\subsection{Graph-based Model: \codegraph{}}
We use CodeGraph4CCDetector \citep{CodeGraph4CCDetector} as the graph based model for our source code clone detection, it is from here on refered to as \codegraph{} model. This model initially is used by its authors on \bcb{} data-set for its classification, and hence we keep the pipeline as it is. However, we trained our own word2vec model \citep{word2vec}, using gensim \citep{gensim_framework} to keep it consistent with respect to the sequence model. We call this model as \codetwovec{} model, which helps to generate the source code token embedding for our source code. We train this \codetwovec{} model using the source tokens which are tokenized by the \codebert{}'s \citep{CodeBERT} tokenizer, which is a Byte Pair encoding tokenizer. This helps in two ways, Firstly, it helps to keep it consistent with the comparison of the sequence model, and secondly it helps to retain the word meanings of the human written source code variable names etc. 

Once we get the tokenized vector format of each graph nodes using the \codetwovec{} model on every node of CPG, we then use the \codegraph{} architecture as it is. Here similar to the sequence model, we pass the code pair sequentially to the \codegraph{} model, which then generates the graph level representation. This representation is the taken to a shallow LSTM layer which helps to perform a binary classification on this graph level representation.

\section{Experimental Design}\label{sec:experimental_design}

Based on our proposed methodology, we conduct experimentation on the following research questions (RQs):

\begin{itemize}
    \item RQ1: Will a graph-based model that leverages both structural and semantic information surpass a sequence-based model in an in-domain static source code analysis?
    \item RQ2: Will a graph-based model trained on multiple source code languages outperform a sequence-based model in cross-lingual generalization and semantic extraction?
    \item RQ3: Will a graph-based model excel over a sequence-based model in the domain adaptation of zero-shot source code clone classification?
\end{itemize}

\subsection{Experiment Data}\label{subsec:experiment_data}

For our experimental setups, we perform clone detection on two publicly available data-sets. The first one is Big Clone Bench (\bcb{}), which is a java language data-set that was originally introduced by \citet{BCB_original}. We used the version of \bcb{} that was filtered according to FA-AST \citep{FA_AST_BCB_filterd}. \bcb{} contains 9,134 java methods, which generate over 2M combinations of clone and non-clone code pairs. The second one is \poolc{}, which consists of over 6M python code snippets that were extracted from hugging face \citep{PoolC}.

We applied various parameters to limit the data-set for the experimentation phase. We restricted the number of lines to be between 5 and 100, the maximum number of characters to be 2000, and the maximum number of nodes in the graph to be 100. The details of how the distribution changed before and after applying the thresholds are given in Appendix \ref{appendix:dataset}. Table \ref{table:dataset_properties} shows the summary of the average counts for the filtered files according to each data-set (\bcb{}, \poolc{}, and their combination, \mixone{}).

\begin{table}[]
    \centering
    \begin{tabular}{cccc}
        \hline \textbf{Attribute} &  \textbf{\bcb{}} & \textbf{\poolc{}} & \textbf{\mixone{}} \\ 
        & (Java) & (Python) & (Java + Python) \\
        \hline 
         Actual File Counts  & 9,126 & 44,950 & - \\
         Filtered File Counts & 2,048 & 17,570 & 19,063 \\
         \hline
         Avg* Lines & 12 & 10 & 10 \\
         Avg* Characters & 450 & 158 & 190 \\
         Avg* Tokens & 200 & 83 & 96 \\
         \hline
         Avg* Nodes & 76 & 67 & 68 \\
         Avg* Leaf Nodes & 36 & 32 & 32 \\
         Avg* AST Edges & 75 & 66 & 67 \\
         Avg* DFG Edges & 15 & 22 & 21 \\
         \hline \multicolumn{4}{l}{*Avg: Average on the filtered files.}
    \end{tabular}
    \caption{Data-set counts of actual and filtered file counts, with their static metrics.}
    \label{table:dataset_properties}
\end{table}

We randomly sampled pairs of clone and non-clone from the filtered files set to form the data-set. Table \ref{table:dataset_pairs} summarizes the data-set pairs according to each data-set.

\begin{table}[]
    \centering
    \begin{tabular}{ccccc}
        \hline \textbf{Dataset} & \textbf{Split} & \textbf{Total pairs} & \textbf{Positive} & \textbf{Negative} \\ 
        \hline 
        & Train & 50,855 & 29,070 & 21,785 \\
        \bcb{}  & Test & 4,000 & 2,000 & 2,000 \\
        & Val & 4,000 & 2,000 & 2,000 \\
        \hline 
        & Train & 50,500 & 25,250 & 25,250 \\
        \poolc{}  & Test & 4,000 & 2,000 & 2,000 \\
        & Val & 4,000 & 2,000 & 2,000 \\
        \hline 
        & Train & 50,000 & 25,000 & 25,000 \\
        \mixone{}  & Test & 4,000 & 2,000 & 2,000 \\
        & Val & 4,000 & 2,000 & 2,000 \\
        \hline \multicolumn{4}{l}{Positive: Clone pairs | Negative: Not a Clone pair.}
    \end{tabular}
    \caption{Data-set sample size. Equally sampled from each of the data-sets.}
    \label{table:dataset_pairs}
\end{table}

\subsection{Experimental Setup}
We chose the state-of-the-art sequence model and graph model, namely \codebert{} \citep{CodeBERT} and \codegraph{} \citep{CodeGraph4CCDetector}, respectively, to conduct various experiments. To answer the research questions, we designed the experiments around them as follows.

\begin{itemize} 
    \item Experiment 1: We train and evaluate Sequence and Graph models independently on each of the data-sets, namely \bcb{} and \poolc{}, to compare their performance within the same domain as baselines.
    \item Experiment 2: We train and evaluate Sequence and Graph models on \mixone{} Data-set, which is a mixture of data from both domains, to examine their cross-domain learning and generalization capabilities. 
    \item Experiment 3: We train Sequence and Graph models on \bcb{} data-set and test them on \poolc{} data-set, and vice versa, to assess their cross-domain zero-shot performance. 
\end{itemize}


\subsection{Model Hyper-parameters}

We use the same machines with Intel® Xeon® Gold 5222 and one Quadro RTX 6000 to train both the sequence and graph models (\codebert{} and \codegraph{}, respectively) in order to maintain a consistent experimentation environment. The maximum batch size that \codebert{} can run on a single RTX 6000 is 16 code pairs, or 32 code snippets per batch. We also set the batch size of \codegraph{} to the same value. The other hyper-parameters used for training these models are given in Appendix \ref{appendix:training}.

\section{Results}\label{sec:results}

\subsection{Experiment 1}
We train the sequence and graph models (\codebert{} and \codegraph{}, respectively) on two datasets: \bcb{} and \poolc{}. This leads to four model trainings and evaluations, as shown in Table \ref{table:exp_1} on page \pageref{table:exp_1}. We select the best-performing epochs for each model, which are the 3\textsuperscript{rd} epoch for \codebert{} and the 2\textsuperscript{nd} epoch for \codegraph{}. We find that \codegraph{} consistently outperforms \codebert{} on both datasets, demonstrating that \codegraph{} has a better learning capability on the source code than \codebert{} under limited data and constrained environment conditions. We highlight statistically significant experimental results in the tables based on  bootstrap testing \citep{bootstrapTesting} with p value below 0.05 for statistical significance, which compares \codebert{} and \codegraph{}.

\begin{mdframed}
    \textbf{Answer to RQ1:} Baseline on the \bcb{} and \poolc{} data-sets, suggests that the graph based model outperforms the sequence based model. This suggests that the graph model can better capture the structural and semantic information of the source code than the sequence model.
\end{mdframed}

\begin{table}[H]
    \centering
    \begin{tabular}{ccccccc}
        \hline 
        \multirow{2}{*}{\textbf{\shortstack{Model\\Name}}} & \multirow{2}{*}{\textbf{\shortstack{Train \& Eval \\Dataset}}} & 
            \multicolumn{4}{c}{\textbf{Metrics}} \\ 
            \cline{3-6} & & \textbf{\textit{A}} & \textbf{\textit{P}} & \textbf{\textit{R}} & \textbf{\textit{F1}} \\
        \hline
        \codebert{} & \multirow{2}{*}{\bcb{}} & 97.62 & 97.63 & 97.62 & 97.62 \\
        \codegraph{} & & \textbf{98.88*} & \textbf{98.88*} & \textbf{98.88*} & \textbf{98.87*} \\
        \hline
        \codebert{} & \multirow{2}{*}{\poolc{}} & 81.82 & 83.92 & 81.82 &  81.54 \\
        \codegraph{} & & \textbf{84.00*} & \textbf{84.86} & \textbf{84.00*} & \textbf{83.90*} \\
        \hline 
        \multicolumn{4}{l}{A: accuracy, P: precision, R: recall, F1: F-score} \\
        \multicolumn{4}{l}{Bold is best value, * is statistically significant.}
    \end{tabular}
    \caption{Experiment 1 | Performance of Graph-Based and Sequence-Based Models on \bcb{} and \poolc{} Data-Sets.}
    \label{table:exp_1}
\end{table}

\subsection{Experiment 2}

We use the same model architectures from Experiment 1, but we train them on a cross-lingual data-set (\mixone{}) that combines both the \bcb{} and \poolc{} data sets. We then evaluate these models on the \mixone{} data-set as well as the individual \bcb{} and \poolc{} data-sets. The results are shown in Table \ref{table:exp_2} on page \pageref{table:exp_2}.

The evaluation results on the \mixone{} dataset for both \codebert{} and \codegraph{} are intermediate between the single-language models trained in Experiment 1. This is further confirmed by the evaluation results on the individual \bcb{} and \poolc{} data-sets, where we observe that cross-lingual training improves the performance of \codegraph{} on both data-sets, from 83.90 to 87.64 on the \poolc{} data-set and from 98.87 to 99.42 on the \bcb{} data-set, indicating that \codegraph{} generalizes better on the source code with cross-lingual training. On the other hand, we observe that cross-lingual training does not improve the performance of \codebert{} as much as \codegraph{}, decreasing it by 0.55 on the \bcb{} data-set and increasing it by only 0.19 on the \poolc{} data-set.

\begin{mdframed}
    \textbf{Answer to RQ2:} The results on the cross-lingual setting of \codebert{} and \codegraph{} models, i.e. trained on \mixone{} data-set, demonstrate that \codegraph{} is a more generalized model than \codebert{} as evidenced by the improvement in the performance of \codegraph{} especially on \poolc{} data-set, whereas we observe a decline in the performance of \codebert{} on \bcb{} data-set and marignal improvement on \poolc{} dataset. This implies that graph models are more adaptable for cross-lingual source code analysis.
\end{mdframed}

\begin{table}[H]
    \centering
    \begin{tabular}{ccccccc}
        \hline
        \multirow{2}{*}{\textbf{\shortstack{Model\\Name}}} & \multirow{2}{*}{\textbf{\shortstack{Train\\Dataset}}} & \multirow{2}{*}{\textbf{\shortstack{Eval\\Dataset}}} & 
            \multicolumn{4}{c}{\textbf{Metrics}} \\ 
            \cline{4-7} & & & \textbf{\textit{A}} & \textbf{\textit{P}} & \textbf{\textit{R}} & \textbf{\textit{F1}} \\
        \hline 
        \codebert{} & \multirow{2}{*}{\mixone{}} &  \multirow{2}{*}{\mixone{}} & 90.35 & 90.51 & 90.35 & 90.34  \\
        \codegraph{} & & & \textbf{93.65*} & \textbf{93.77*} & \textbf{93.65*} & \textbf{93.65*} \\
        \hline
        \codebert{} & \multirow{2}{*}{\mixone{}} &  \multirow{2}{*}{\bcb{}} & 97.08 & 97.11 & 97.08 & 97.07  \\
        \codegraph{} & & & \textbf{99.42*} & \textbf{99.43*} & \textbf{99.42*} & \textbf{99.42*} \\
        \hline
        \codebert{} & \multirow{2}{*}{\mixone{}} &  \multirow{2}{*}{\poolc{}} & 81.82 & 82.53 & 81.82 & 81.73  \\
        \codegraph{} & & & \textbf{\underline{87.68}*} & \textbf{\underline{88.13}*} & \textbf{\underline{87.68}*} & \textbf{\underline{87.64}*} \\
        \hline
        \multicolumn{6}{l}{A: accuracy, P: precision, R: recall, F1: F-score} \\
        \multicolumn{6}{l}{Bold is best value, * is statistically significant} \\
        \multicolumn{6}{l}{Underlined is statistically significant \& better w.r.t Experiment 1.}
    \end{tabular}
    \caption{Experiment 2 | Performance of Graph-Based and Sequence-Based Models on \mixone{} Data-Set.}
    \label{table:exp_2}
\end{table}

\subsection{Experiment 3}

We test the domain adaptation of the pre-trained models from Experiment 1, i.e., \codebert{} and \codegraph{}, on a different source code language than the one they were trained on. For instance, we evaluate \codebert{} trained on \bcb{} on \poolc{}, and vice versa. We repeat the same procedure with \codegraph{} without changing the environment. The results of this experiment are shown in Table \ref{table:exp_3} on page \pageref{table:exp_3}.

This experiment simulates the domain adaptation from Python source code to Java source code and vice versa. The results show that \codebert{} performs very poorly on a different domain, with F1 scores of 33.71 and 36.56 for \poolc{} and \bcb{} evaluation, respectively. This indicates that the model has over-fitted on the domain and cannot generalize well to a new domain. We observe the same trend with other epochs. On the other hand, \codegraph{} performs much better than \codebert{} on a different domain, with F1 scores of 53.67 and 46.44 for \poolc{} and \bcb{} evaluation, respectively. This demonstrates that \codegraph{} has a better domain adaptation capability than \codebert{} in a zero-shot learning setting, although it does not achieve state-of-the-art performance. This suggests that representing source code as a graph rather than a sequence is a promising direction for future research.

\begin{mdframed}
    \textbf{Answer to RQ3:} The results on the domain adaptation task show that \codegraph{} outperforms \codebert{} in adapting to a new source code language domain without any labeled data for that domain during training. This indicates that graph-based model has an advantage over sequence-based model in the domain adaptation of zero-shot source code clone classification task.
\end{mdframed}

\begin{table}[H]
    \centering
    \begin{tabular}{ccccccc}
        \hline
        \multirow{2}{*}{\textbf{\shortstack{Model\\Name}}} & \multirow{2}{*}{\textbf{\shortstack{Train\\Dataset}}} & \multirow{2}{*}{\textbf{\shortstack{Eval\\Dataset}}} & 
            \multicolumn{4}{c}{\textbf{Metrics}} \\ 
            \cline{4-7} & & & \textbf{\textit{A}} & \textbf{\textit{P}} & \textbf{\textit{R}} & \textbf{\textit{F1}} \\
        \hline
        \codebert{} & \multirow{2}{*}{\bcb{}} & \multirow{2}{*}{\poolc{}} & 50.05 & 53.58 & 50.05 & 33.71 \\
        \codegraph{} & & & \textbf{53.67} & \textbf{53.68} & \textbf{53.68*} & \textbf{53.67*} \\
        \hline 
        \codebert{} & \multirow{2}{*}{\poolc{}} & \multirow{2}{*}{\bcb{}} & 48.95 & 45.20 & 48.95 & 36.56 \\
        \codegraph{} & & & \textbf{54.88*} & \textbf{63.16*} & \textbf{54.88*} & \textbf{46.44*}  \\
        \hline 
        \multicolumn{4}{l}{A: accuracy, P: precision, R: recall, F1: F-score} \\
        \multicolumn{4}{l}{Bold is best value, * is statistically significant.}
    \end{tabular}
    \caption{Experiment 3 | Performance of Graph-Based and Sequence-Based Models on Cross-Domain Zero-Shot Evaluation.}
    \label{table:exp_3}
\end{table}

\subsection{Discussion}
We analyze the false predictions made by both the models, \codebert{} and \codegraph{}, and find that most of them are false positives, especially from CodeBERT. When we examine these examples from \codebert{}, we notice that the model predicts them as false positives with high confidence, whereas the \codegraph{} model either predicts them as true negatives or false positives with low confidence. This indicates that adjusting the classification threshold for \codegraph{} could improve its overall performance. However, for \codebert{}, we observe that the model is confused by the very similar keywords in the code pair. We provide a detailed analysis of this in Appendix \ref{appendix:results:FPanalysis}.

We also analyze the false negatives for \codegraph{} on the \poolc{} data-set, which are the most frequent among all the models and data-sets. We find that these false negatives are mainly due to the large size differences between the code pairs in the \poolc{} data-set. The examples we inspect are clones of type IV, but they have one code snippet much longer than the other. This makes it difficult for \codegraph{} to recognize them as clones and it predicts them as non-clones instead. We provide some detailed explanation and examples of these false negatives in Appendix \ref{appendix:results:FNanalysis}.

\section{Limitations and Future Research}\label{sec:limitations_n_future_research}

\subsection{Limitations}\label{subsec:limitations}

We note the following limitations and concerns in the study:

\begin{itemize}

    \item The experimental setup data-set size is drastically reduced in order to time-bound the experiments for this research project. Majorly, there are 2 cuts in the data-set size, Firstly, source code files are filtered to only those which have less than or equal to 100 nodes. Secondly, the sample size of the data-set clone and no clone pair is restricted to 50K data-points only. Refer the Appendix \ref{appendix:dataset} to check the filtering criterion and Section \ref{subsec:experiment_data} to understand the data-point split samples.

    \item The \bcb{} data-set was significantly reduced after applying the thresholding criterion, resulting in only 2K files out of the original 9K. This led to a over-sampled distribution of the training pairs, which consisted of 50K data-points. As shown in the Results Section \ref{sec:results}, this caused the \bcb{} data-set model to over-fit the data, unlike the \poolc{} data-set model.
    
    \item A potential limitation of this study is the discrepancy in the number of trainable model parameters between the sequence model (\codebert{}) and the graph model (\codegraph{}). The sequence model has 125M parameters, which is 125 times more than the graph model’s 1.1M parameters. This could raise the question of whether the graph model’s superior performance is due to its inherent advantages or its lower complexity. However, this also suggests that there is room for further improvement on the graph model by increasing its number of parameters. 
    
    
\end{itemize}

\subsection{Future Research}\label{subsec:future_research}

Some possible directions for the future research based on the limitations are as follows:

\begin{itemize}
    \item To evaluate the impact of data-set size on the performance of the models, future research could use the complete and more diverse data-set that include source code files with more than 100 nodes and data-set samples itself going upwards of a million samples. This would help to test the generalizability and robustness of the models across different domains and languages at a larger scale.

    \item To help reduce over fitting problem on the Java data-set (i.e. \bcb{} data-set), future research could use samples from other data-sets, like CodeForces \citep{codeForces}, Google Code Jam \citep{GCJ}, which can yield in more diversified data-set for Java language.

    \item To explore the potential of the graph model (\codegraph{}), future research could increase its number of trainable parameters and compare its performance with the sequence model (\codebert{}) under the same complexity level. This would help to determine whether the bigger graph model would still have inherent advantages over the sequence model or not. conversely, one can reduce the parameters on sequence model and check its impact. 

    
    \item \poolc{} data-set false Negatives are majorly due to the code size length differences. This can be solved if trained with longer snippet of codes. 

    \item Train a mixture model on various source code languages, not just limiting to two, such as JavaScript, SQL, HTML, etc., and evaluate its generalization ability on different domains together. Moreover, cross-domain example pairs could be generated from Code Forces \citep{codeForces}, which is an online platform for competitive programming that supports multiple languages.

\end{itemize}

\section{Conclusion}\label{sec:conclusion}

In this paper, we have shown that graph-based methods are superior to sequence-based methods for source code clone detection. We have used the state-of-the-art models \codebert{} \citep{CodeBERT} and \codegraph{} \citep{CodeGraph4CCDetector} to conduct various experiments on two benchmark data-sets: \bcb{} \citep{BCB_original} and \poolc{} \citep{PoolC}. We have demonstrated that graph models can better capture the structural and semantic information of the source code than sequence models in a series of 3 experimental setups, and that they can generalize better across different source code languages and domains. We have also provided efficient and scalable code for generating standard CPG representations of source code, along with the re-implemented code for the sequence and graph-based models. Our work has important implications for future research on source code analysis, as it suggests that representing source code as a graph rather than a sequence is a promising direction for enhancing the performance and generalization of static source code analysis models.

\section{Acknowledgements}
We are deeply grateful to our supervisor, \href{https://scholar.google.com/citations?hl=en&user=WMYcG5EAAAAJ}{Dr. Julia Ive}, for her constant guidance, support, and feedback throughout this research project. She has inspired and motivated us with her expertise and experience, and we have learned a lot from her. We also thank Vishal Yadav, Mashhood Alam, and other peers as well as the reviewers for their valuable comments and suggestions that enhanced the quality of our paper. Moreover, we acknowledge the Queen Mary University and organizations that supported this work, as well as the authors of the data-sets and models that we used in our experiments. Lastly, we thank our families for their love and encouragement, and the almighty for his blessings and grace.

\section{Data Availability}

We have made the data and source code that we used in this paper publicly accessible. Source code is available for replication at: \url{https://github.com/Ataago-AI/clone-detection}, and the filtered data-sets can be downloaded from here: \url{https://drive.google.com/drive/folders/1phx8k_JB8HC_HW3nhZLec9BKjRxNCN2b?usp=drive_link}


\bibliographystyle{agsm} 
\bibliography{references}


\newpage
\onecolumn
\appendices
\newpage

\section{Code Representations}\label{appendix:code_rep}

\subsection{Python Standard Code Property Graphs Example pairs}

\begin{figure}[H]
    \centerline{\includegraphics[width=\columnwidth]{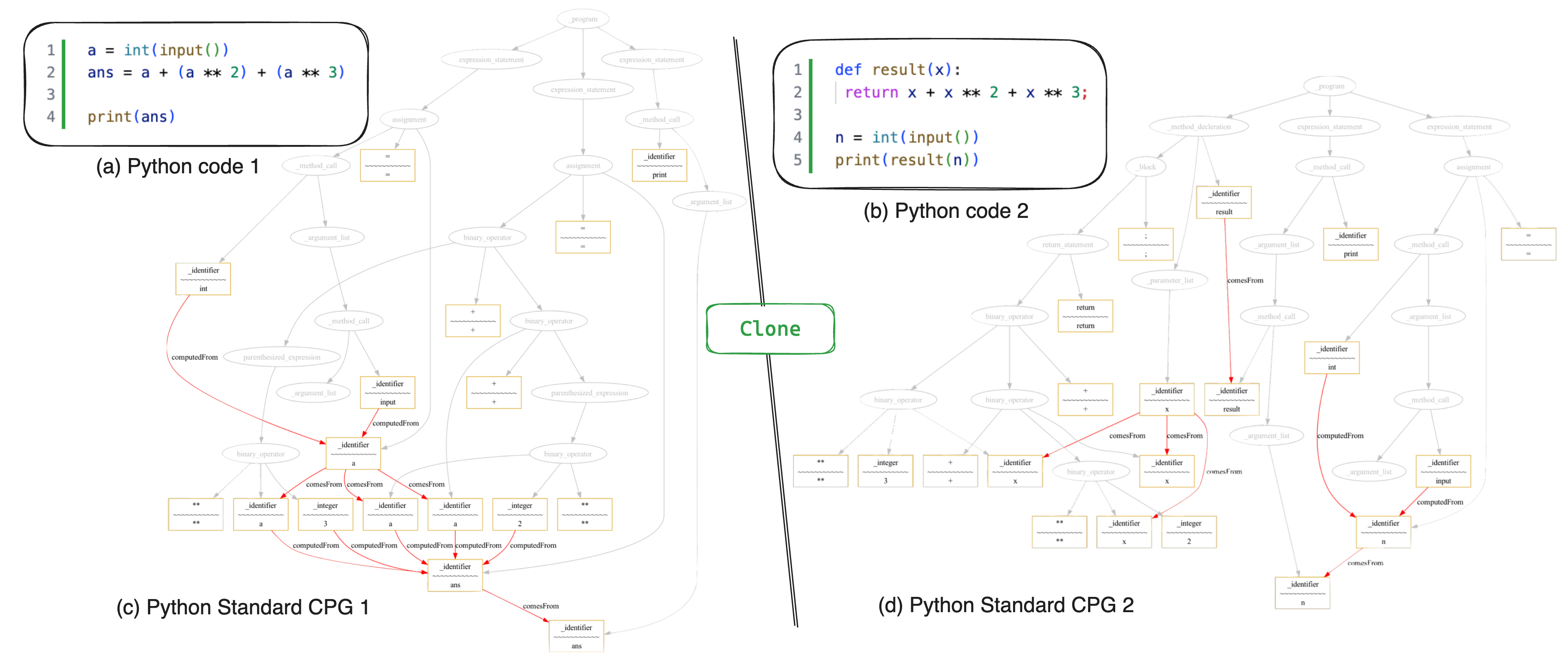}}
    \caption{An example of python code clone pairs with its Standard Code Property Graphs. (a) \& (b) are the source codes, and (c) \& (d) are its respective Standard Code Property Graphs.}
    \label{fig:code_representation_pair_python_001}
\end{figure}
\begin{figure}[H]
    \centerline{\includegraphics[width=\columnwidth]{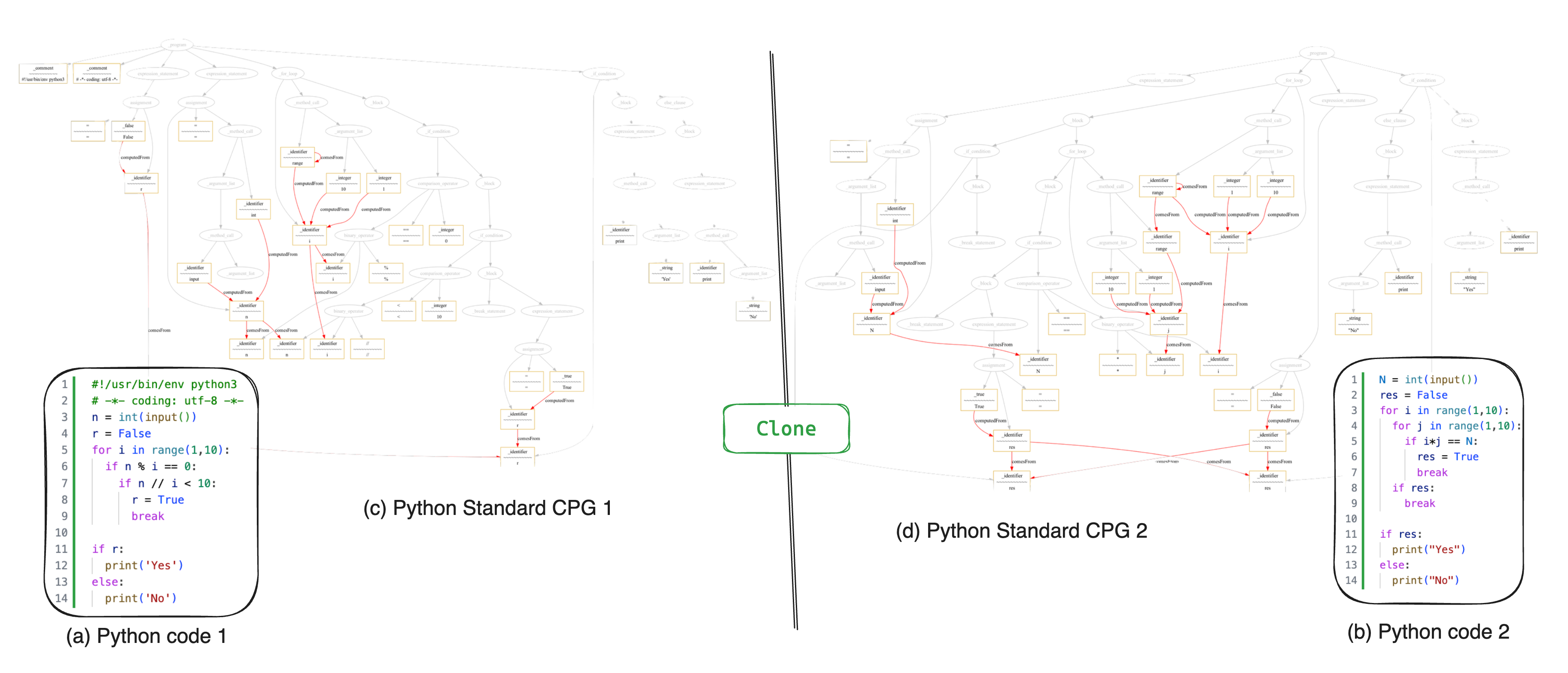}}
    \caption{An example of python code clone pairs with its Standard Code Property Graphs. (a) \& (b) are the source codes, and (c) \& (d) are its respective Standard Code Property Graphs.}
    \label{fig:code_representation_pair_python_002}
\end{figure}

\newpage
\subsection{Java Standard Code Property Graphs Example pairs}
\begin{figure}[H]
    \centerline{\includegraphics[width=\columnwidth]{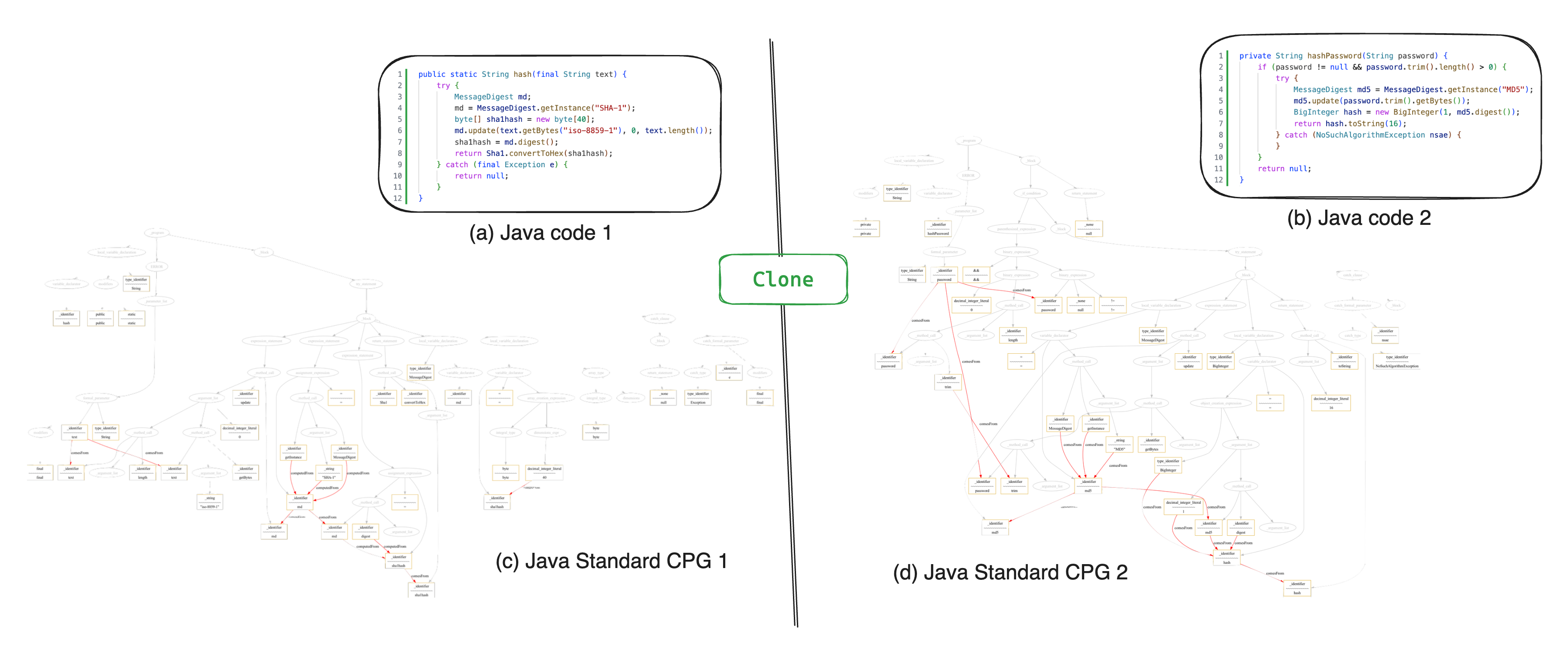}}
    \caption{An example of Java code clone pairs with its Standard Code Property Graphs. (a) \& (b) are the source codes, and (c) \& (d) are its respective Standard Code Property Graphs.}
    \label{fig:code_representation_pair_java_001}
\end{figure}
\begin{figure}[H]
    \centerline{\includegraphics[width=\columnwidth]{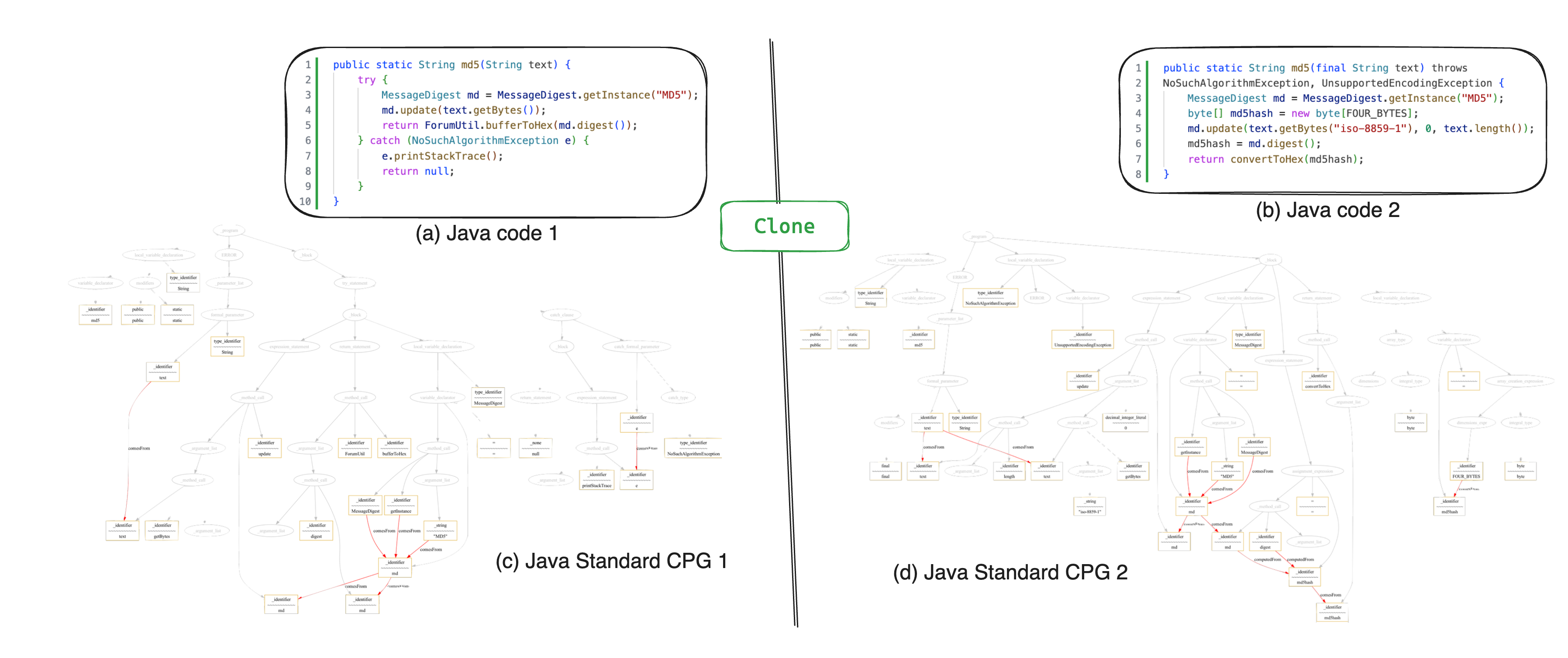}}
    \caption{An example of Java code clone pairs with its Standard Code Property Graphs. (a) \& (b) are the source codes, and (c) \& (d) are its respective Standard Code Property Graphs.}
    \label{fig:code_representation_pair_java_002}
\end{figure}


\newpage
\section{Data-set}\label{appendix:dataset}

Data set is filtered based on various parameters like number of lines, number of characters, and number of nodes. Given below are the charts how the data looks before and after filtering.

\subsection{Java Data : BCB}
\begin{figure}[H]
    \centering
    \begin{subfigure}[b]{0.32\columnwidth}
        \centering
        \includegraphics[width=\columnwidth]{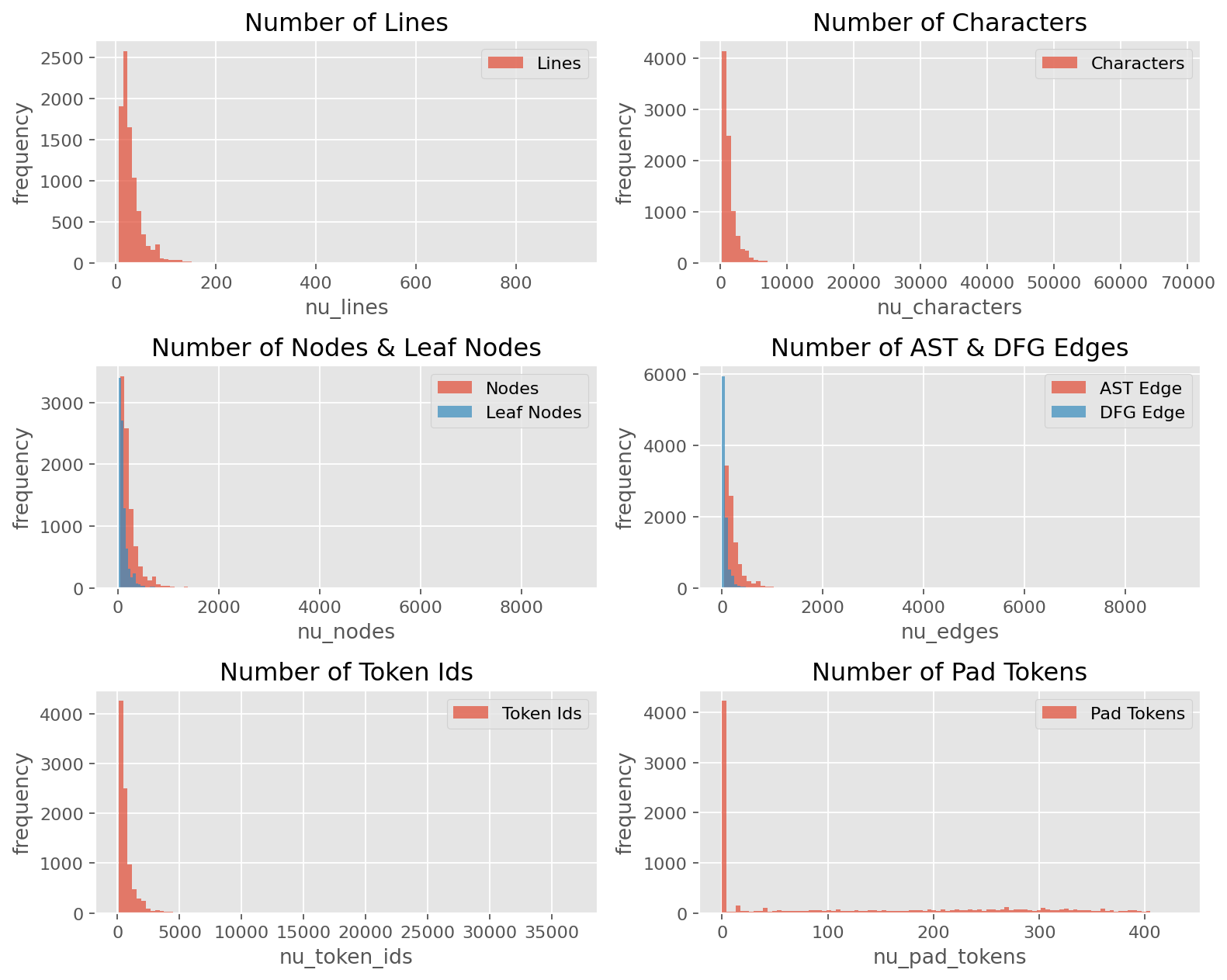}
        \caption{Original}
        \label{fig:bcb_original}
    \end{subfigure}
    \hfill
    \begin{subfigure}[b]{0.32\columnwidth}
        \centering
        \includegraphics[width=\columnwidth]{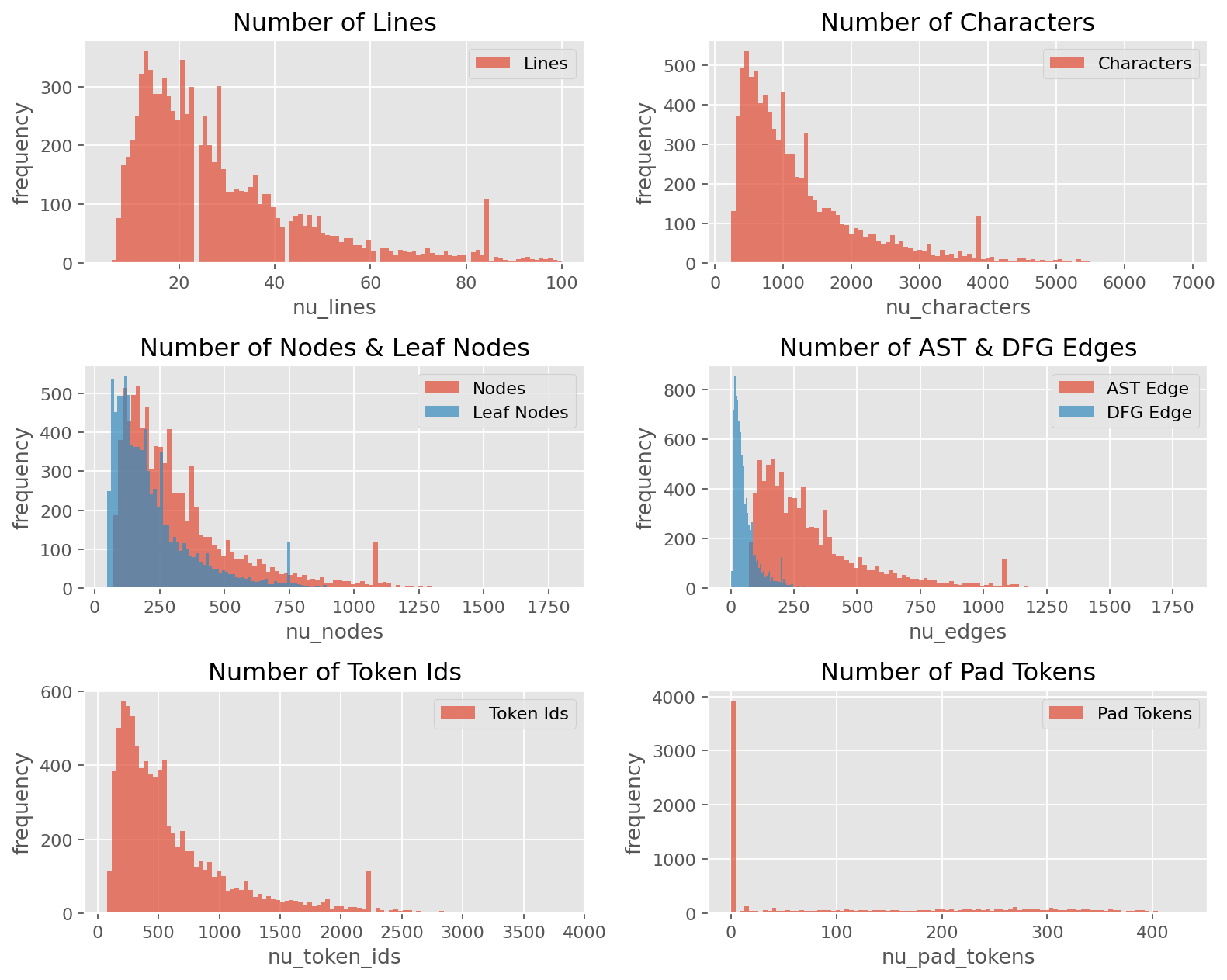}
        \caption{Original (no outliers)}
        \label{fig:bcb_original_no_outliers}
    \end{subfigure}
    \hfill
    \begin{subfigure}[b]{0.33\columnwidth}
        \centering
        \includegraphics[width=\columnwidth]{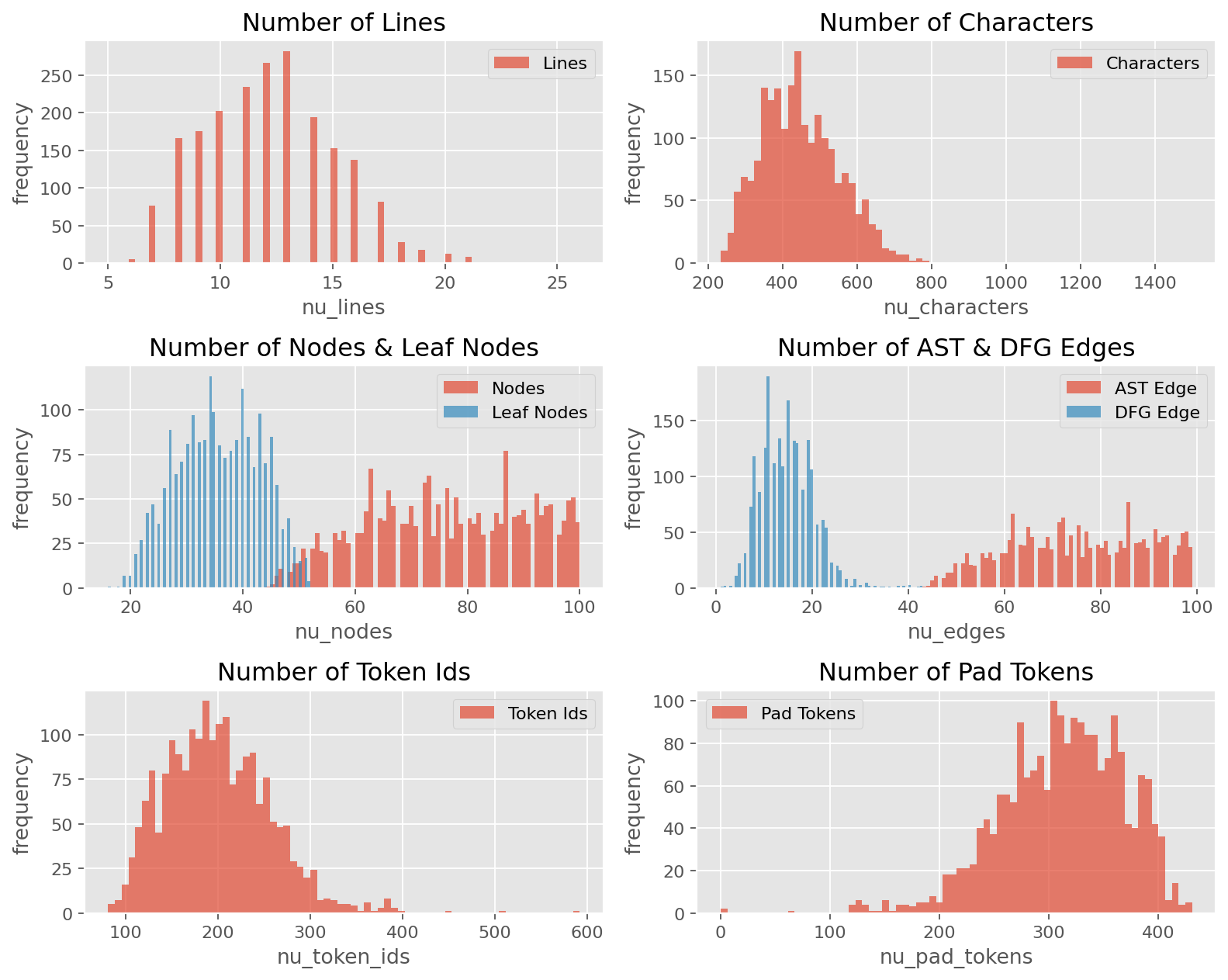}
        \caption{Filtered to Max 100 nodes}
        \label{fig:bcb_filtered}
    \end{subfigure}
    \caption{Python Data : BCB}
    \label{fig:BCB_Dataset}
\end{figure}

Given in Figure \ref{fig:BCB_Dataset} are the original and filtered distributions for Java dataset from BigCloneBench BCB dataset \citep{FA_AST_BCB_filterd}.

\begin{table}[H]
    \centering
        \begin{tabular}{lrrrrrrrr}
\toprule
 & nu\_lines & nu\_characters & nu\_nodes & nu\_leaf\_nodes & nu\_ast\_edges & nu\_dfg\_edges & nu\_token\_ids & nu\_pad\_tokens \\
\midrule
\textbf{count} & 9126.0 & 9126.0 & 9126.0 & 9126.0 & 9126.0 & 9126.0 & 9126.0 & 9126.0 \\
\textbf{mean} & 33.9 & 1579.9 & 247.6 & 121.3 & 246.6 & 74.9 & 787.0 & 113.8 \\
\textbf{std} & 40.0 & 2415.3 & 327.0 & 167.0 & 327.0 & 153.9 & 1270.3 & 133.6 \\
\textbf{min} & 5.0 & 234.0 & 44.0 & 16.0 & 43.0 & 1.0 & 81.0 & 0.0 \\
\textbf{25\%} & 16.0 & 605.2 & 105.0 & 50.0 & 104.0 & 23.0 & 277.0 & 0.0 \\
\textbf{50\%} & 24.0 & 989.0 & 169.0 & 82.0 & 168.0 & 42.0 & 478.0 & 34.0 \\
\textbf{75\%} & 38.0 & 1707.0 & 271.0 & 132.0 & 270.0 & 77.0 & 841.0 & 235.0 \\
\textbf{max} & 917.0 & 68541.0 & 9041.0 & 4811.0 & 9040.0 & 5981.0 & 36823.0 & 431.0 \\
\bottomrule
\end{tabular}

        \caption{BCB original distribution}
        \label{tab:bcb_orignal_dist}
        \vspace{15px}
        \begin{tabular}{lrrrrrrrr}
\toprule
 & nu\_lines & nu\_characters & nu\_nodes & nu\_leaf\_nodes & nu\_ast\_edges & nu\_dfg\_edges & nu\_token\_ids & nu\_pad\_tokens \\
\midrule
\textbf{count} & 2048.0 & 2048.0 & 2048.0 & 2048.0 & 2048.0 & 2048.0 & 2048.0 & 2048.0 \\
\textbf{mean} & 12.2 & 449.9 & 76.3 & 35.7 & 75.3 & 14.8 & 199.7 & 312.4 \\
\textbf{std} & 3.0 & 107.8 & 14.3 & 7.4 & 14.3 & 5.5 & 56.6 & 56.4 \\
\textbf{min} & 5.0 & 234.0 & 44.0 & 16.0 & 43.0 & 1.0 & 81.0 & 0.0 \\
\textbf{25\%} & 10.0 & 369.0 & 65.0 & 30.0 & 64.0 & 11.0 & 157.0 & 274.8 \\
\textbf{50\%} & 12.0 & 440.0 & 76.0 & 35.0 & 75.0 & 15.0 & 195.0 & 317.0 \\
\textbf{75\%} & 14.0 & 520.0 & 89.0 & 41.0 & 88.0 & 19.0 & 237.2 & 355.0 \\
\textbf{max} & 26.0 & 1500.0 & 100.0 & 52.0 & 99.0 & 43.0 & 592.0 & 431.0 \\
\bottomrule
\end{tabular}

        \caption{BCB filtered distribution}
        \label{tab:bcb_filtered_dist}
\end{table}


     
 





  

\newpage
\subsection{Python Data : PoolC}
\begin{figure}[H]
    \centering
    \begin{subfigure}[b]{0.3\columnwidth}
        \centering
        \includegraphics[width=\columnwidth]{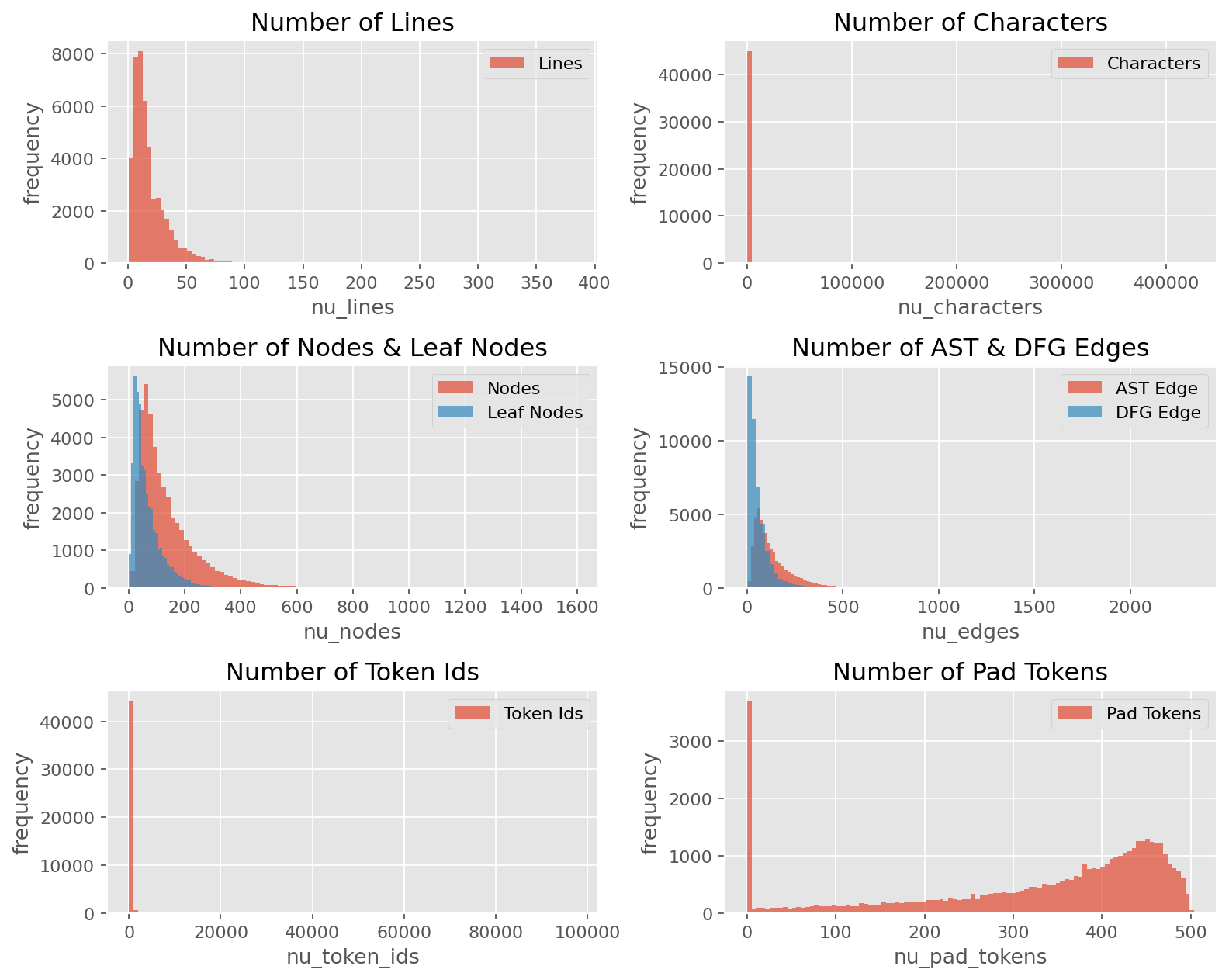}
        \caption{Original}
        \label{fig:poolc_original}
    \end{subfigure}
    \hfill
    \begin{subfigure}[b]{0.3\columnwidth}
        \centering
        \includegraphics[width=\columnwidth]{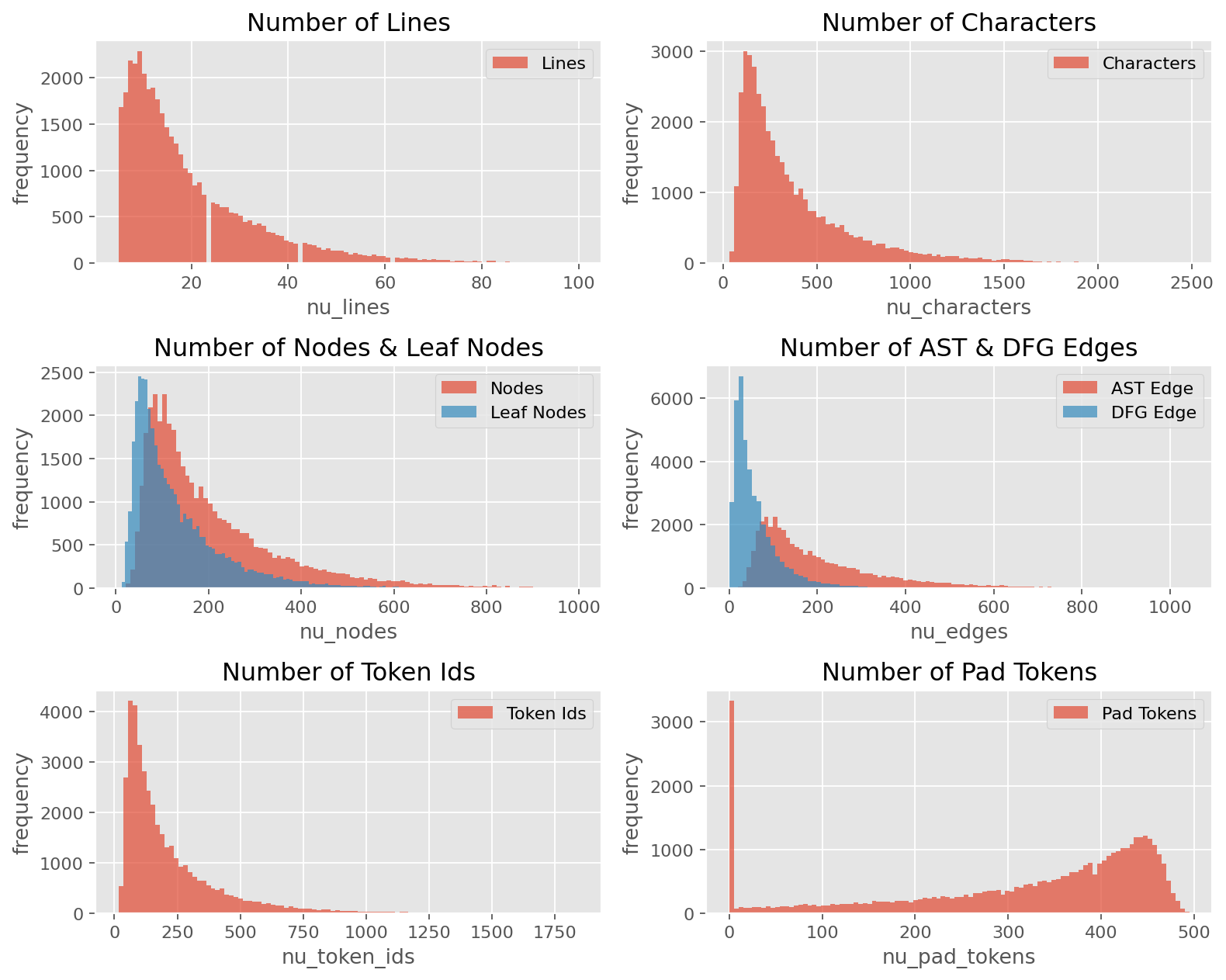}
        \caption{Original (no outliers)}
        \label{fig:poolc_original_no_outliers}
    \end{subfigure}
    \hfill
    \begin{subfigure}[b]{0.3\columnwidth}
        \centering
        \includegraphics[width=\columnwidth]{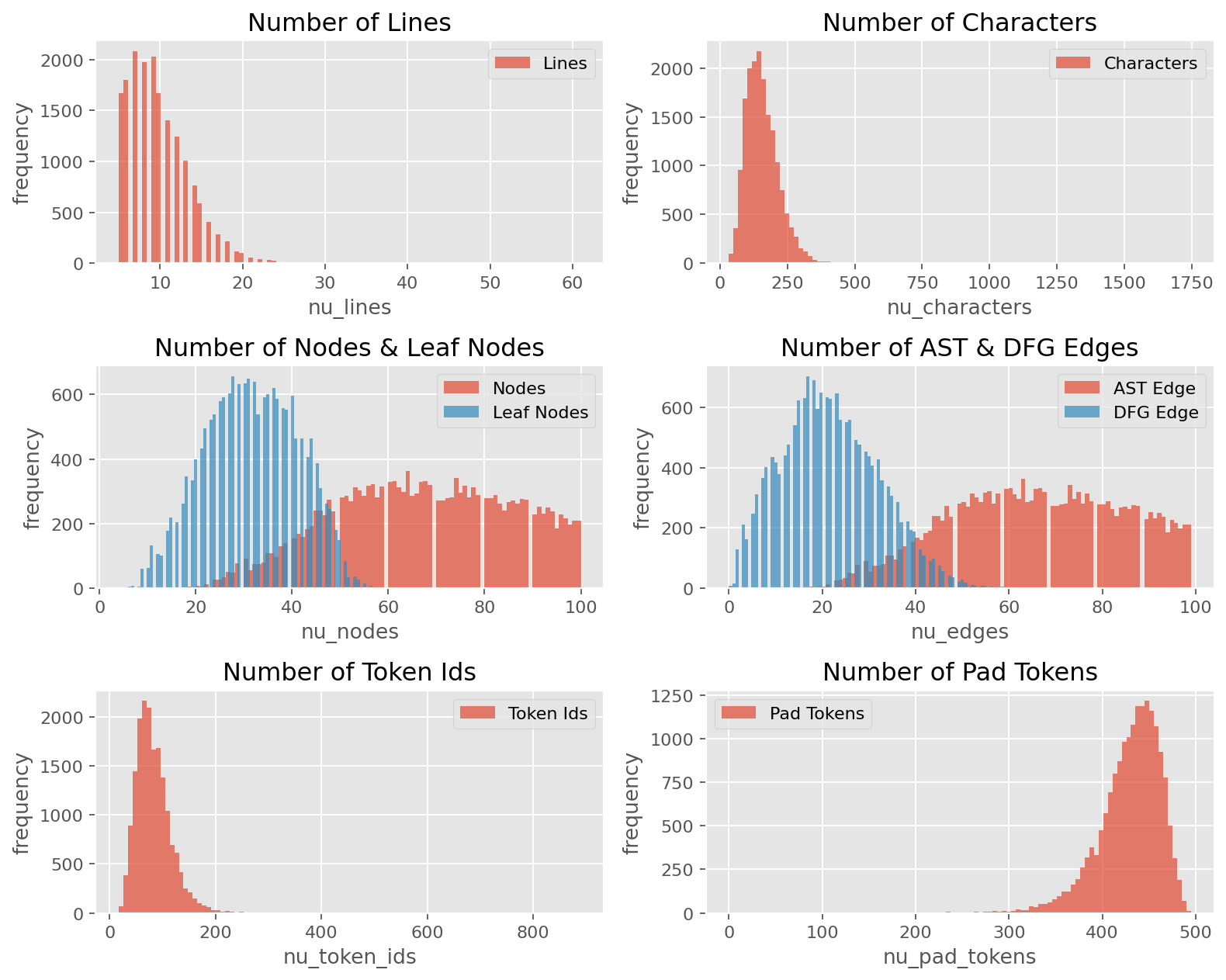}
        \caption{Filtered to Max 100 nodes}
        \label{fig:poolc_filtered}
    \end{subfigure}
    \caption{Python Data : PoolC}
    \label{fig:PoolC_Dataset}
\end{figure}

Given in Figure \ref{fig:PoolC_Dataset} are the original \ref{fig:poolc_original} and filtered \ref{fig:poolc_filtered} distributions for Python dataset from PoolC dataset \citep{PoolC}.
  

\begin{table}[H]
    \centering
        \begin{tabular}{lrrrrrrrr}
\toprule
 & nu\_lines & nu\_characters & nu\_nodes & nu\_leaf\_nodes & nu\_ast\_edges & nu\_dfg\_edges & nu\_token\_ids & nu\_pad\_tokens \\
\midrule
\textbf{count} & 44950.0 & 44950.0 & 44950.0 & 44950.0 & 44950.0 & 44950.0 & 44950.0 & 44950.0 \\
\textbf{mean} & 19.1 & 392.4 & 141.5 & 69.8 & 140.5 & 58.6 & 213.6 & 326.6 \\
\textbf{std} & 17.5 & 2061.9 & 118.9 & 61.7 & 118.9 & 68.2 & 538.9 & 147.5 \\
\textbf{min} & 1.0 & 16.0 & 6.0 & 2.0 & 5.0 & 0.0 & 8.0 & 0.0 \\
\textbf{25\%} & 8.0 & 137.0 & 63.0 & 29.0 & 62.0 & 19.0 & 71.0 & 251.0 \\
\textbf{50\%} & 14.0 & 248.0 & 105.0 & 51.0 & 104.0 & 38.0 & 132.0 & 380.0 \\
\textbf{75\%} & 24.0 & 475.0 & 182.0 & 90.0 & 181.0 & 74.0 & 261.0 & 441.0 \\
\textbf{max} & 384.0 & 426657.0 & 1596.0 & 846.0 & 1595.0 & 2335.0 & 97574.0 & 504.0 \\
\bottomrule
\end{tabular}

        \caption{Poolc original distribution}
        \label{tab:poolc_orignal_dist}
        \vspace{15px}
        \begin{tabular}{lrrrrrrrr}
\toprule
 & nu\_lines & nu\_characters & nu\_nodes & nu\_leaf\_nodes & nu\_ast\_edges & nu\_dfg\_edges & nu\_token\_ids & nu\_pad\_tokens \\
\midrule
\textbf{count} & 17570.0 & 17570.0 & 17570.0 & 17570.0 & 17570.0 & 17570.0 & 17570.0 & 17570.0 \\
\textbf{mean} & 9.8 & 158.5 & 67.2 & 31.6 & 66.2 & 21.8 & 83.2 & 428.8 \\
\textbf{std} & 3.8 & 68.0 & 18.5 & 9.6 & 18.5 & 10.4 & 36.5 & 36.1 \\
\textbf{min} & 5.0 & 33.0 & 9.0 & 4.0 & 8.0 & 0.0 & 17.0 & 0.0 \\
\textbf{25\%} & 7.0 & 113.0 & 53.0 & 24.0 & 52.0 & 14.0 & 59.0 & 412.0 \\
\textbf{50\%} & 9.0 & 149.0 & 67.0 & 32.0 & 66.0 & 21.0 & 77.0 & 435.0 \\
\textbf{75\%} & 12.0 & 193.0 & 82.0 & 39.0 & 81.0 & 29.0 & 100.0 & 453.0 \\
\textbf{max} & 61.0 & 1748.0 & 100.0 & 69.0 & 99.0 & 69.0 & 890.0 & 495.0 \\
\bottomrule
\end{tabular}

        \caption{Poolc filtered distribution}
        \label{tab:poolc_filtered_dist}
\end{table}

\newpage
\subsection{Java and Python Data : mix 1}
\begin{figure}[H]
    \centering
    \begin{subfigure}[b]{0.3\columnwidth}
        \centering
        \includegraphics[width=\columnwidth]{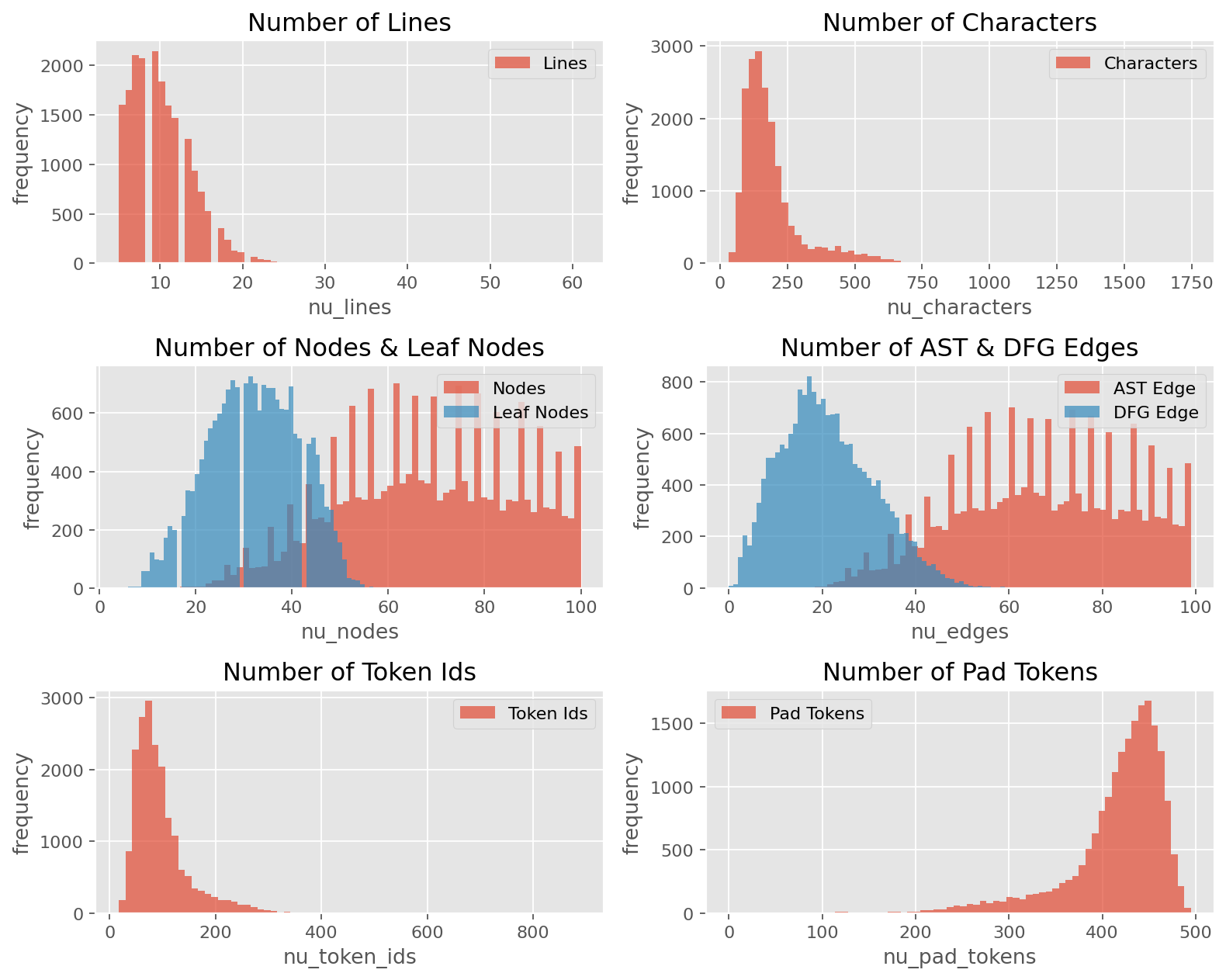}
        \caption{Filtered to Max 100 nodes}
        \label{fig:mix_1_filtered}
    \end{subfigure}
    \caption{Data : Mix 1}
    \label{fig:Mix_1_Dataset}
\end{figure}

Given in Figure \ref{fig:Mix_1_Dataset} we have the distribution for the mixture of BCB and PoolC dataset. This dataset is made by randomly sampling the filtered datasets of BCB and PoolC examples from each of train, valid, and test splits. This results in total 19K files for source code from both Java and Python together, which results in total 25K Java and 25K python labelled pairs. 
  
\begin{table}[H]
    \centering
        \begin{tabular}{lrrrrrrrr}
\toprule
 & nu\_lines & nu\_characters & nu\_nodes & nu\_leaf\_nodes & nu\_ast\_edges & nu\_dfg\_edges & nu\_token\_ids & nu\_pad\_tokens \\
\midrule
\textbf{count} & 19063.0 & 19063.0 & 19063.0 & 19063.0 & 19063.0 & 19063.0 & 19063.0 & 19063.0 \\
\textbf{mean} & 10.1 & 189.7 & 68.2 & 32.0 & 67.2 & 21.0 & 95.7 & 416.4 \\
\textbf{std} & 3.8 & 116.3 & 18.2 & 9.5 & 18.2 & 10.2 & 53.3 & 53.0 \\
\textbf{min} & 5.0 & 33.0 & 9.0 & 4.0 & 8.0 & 0.0 & 17.0 & 0.0 \\
\textbf{25\%} & 7.0 & 117.0 & 54.0 & 25.0 & 53.0 & 14.0 & 61.0 & 400.0 \\
\textbf{50\%} & 9.0 & 157.0 & 68.0 & 32.0 & 67.0 & 20.0 & 82.0 & 430.0 \\
\textbf{75\%} & 12.0 & 214.0 & 83.0 & 39.0 & 82.0 & 28.0 & 112.0 & 451.0 \\
\textbf{max} & 61.0 & 1748.0 & 100.0 & 69.0 & 99.0 & 69.0 & 890.0 & 495.0 \\
\bottomrule
\end{tabular}

        \caption{Mix 1 filtered distribution}
        \label{tab:mix_1_filtered_dist}
\end{table}

\newpage
\section{Training}\label{appendix:training}

\subsection{Model hyper-parameters}

\begin{table}[H]
    \centering
    \begin{tabular}{ccc}
        \hline
        \textbf{Parameter} & \textbf{\codebert{}} & \textbf{\codegraph{}} \\
        \hline 
        \textbf{BATCH\_SIZE}        & 16                & 16 \\
        \textbf{LEARNING\_RATE}     & 5e-05             & 1e-03 \\
        \textbf{OPTIMIZER}          & AdamW             & Adam \\
        \textbf{SCHEDULER}          & OneCylceLR        & NA \\
        \textbf{LOSS\_FUNCTION}     & CrossEntropy      & FocalLoss \\
        \hline 
        \multicolumn{3}{l}{}
    \end{tabular}
    \caption{Hyper-parameters of Sequence \codebert{} and Graph \codegraph{} models.}
    \label{table:hyperparameters}
\end{table}

\subsection{\codebert{} : Sequence Model}


\begin{figure}[H]
    \centering
    \begin{subfigure}[b]{\columnwidth}
        \centering
        \includegraphics[width=\columnwidth]{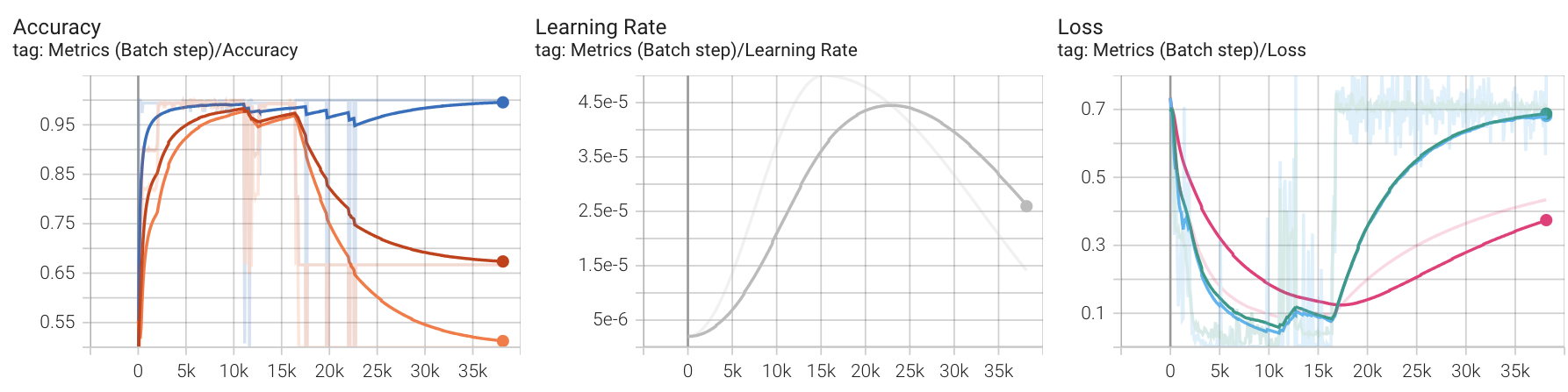}
        \caption{\bcb{} data-set.}
        \label{fig:codebert_bcb}
    \end{subfigure}
    \hfill
    \begin{subfigure}[b]{\columnwidth}
        \centering
        \includegraphics[width=\columnwidth]{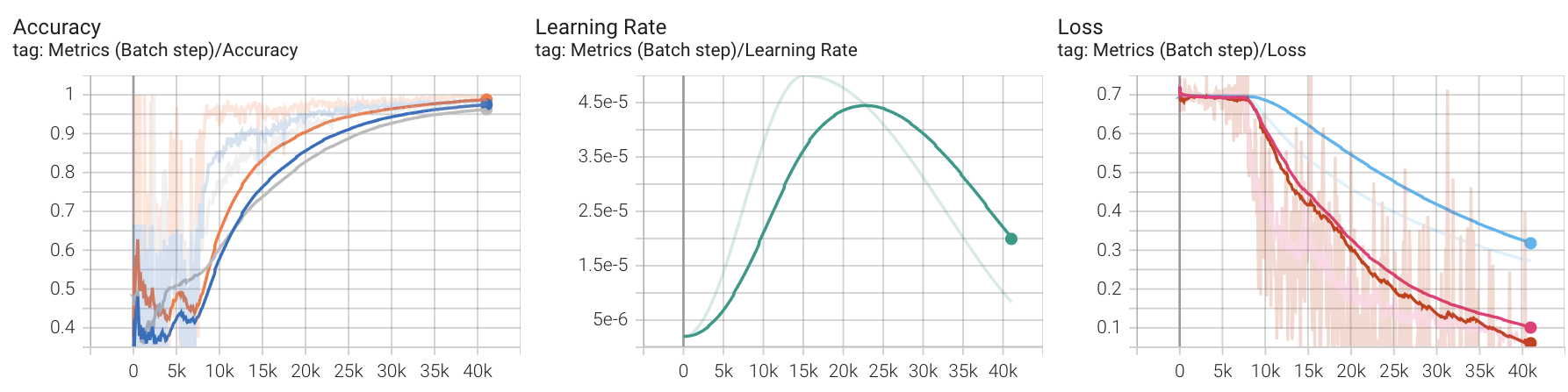}
        \caption{\poolc{} data-set.}
        \label{fig:codebert_poolc}
    \end{subfigure}
    \hfill
    \begin{subfigure}[b]{\columnwidth}
        \centering
        \includegraphics[width=\columnwidth]{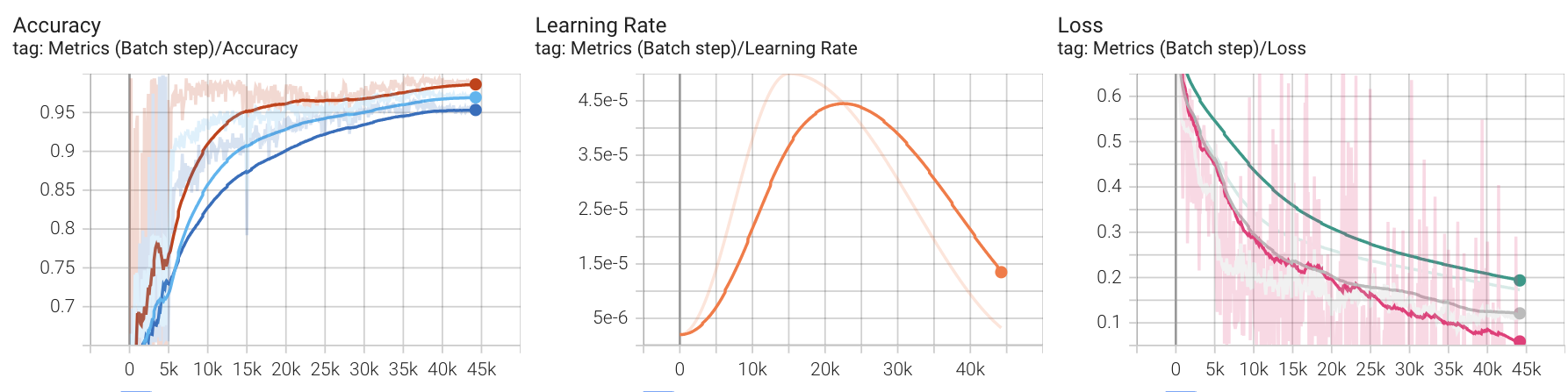}
        \caption{\mixone{} data-set.}
        \label{fig:codebert_mix_1}
    \end{subfigure}
    \caption{Training curves for \codebert{} model}
    \label{fig:training_curves_codebert}
\end{figure}

\subsection{\codegraph{} : Graph Model}
\begin{figure}[H]
    \centering
    \begin{subfigure}[b]{\columnwidth}
        \centering
        \includegraphics[width=\columnwidth]{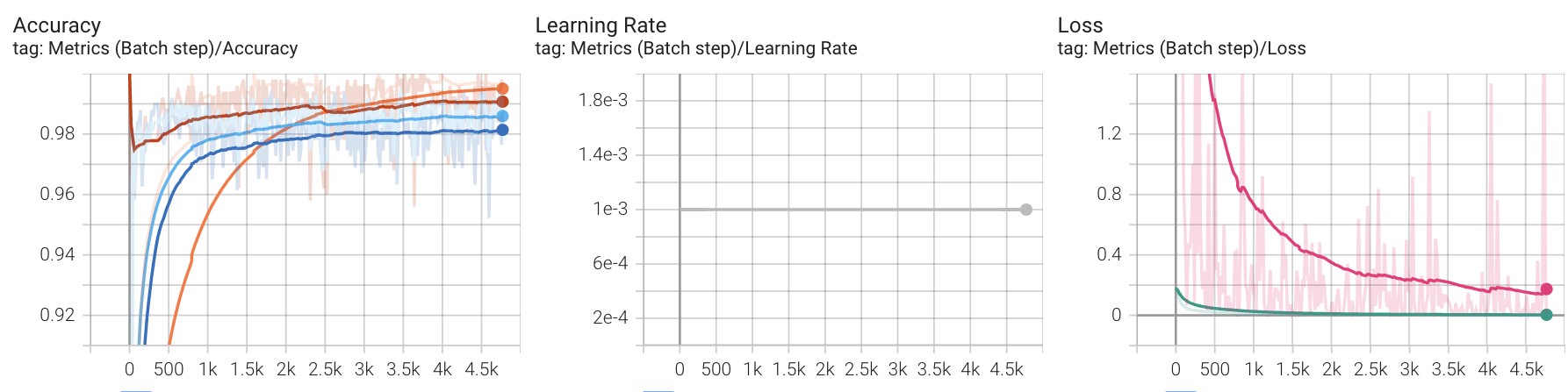}
        \caption{\bcb{} data-set.}
        \label{fig:codegraph_bcb}
    \end{subfigure}
    \hfill
    \begin{subfigure}[b]{\columnwidth}
        \centering
        \includegraphics[width=\columnwidth]{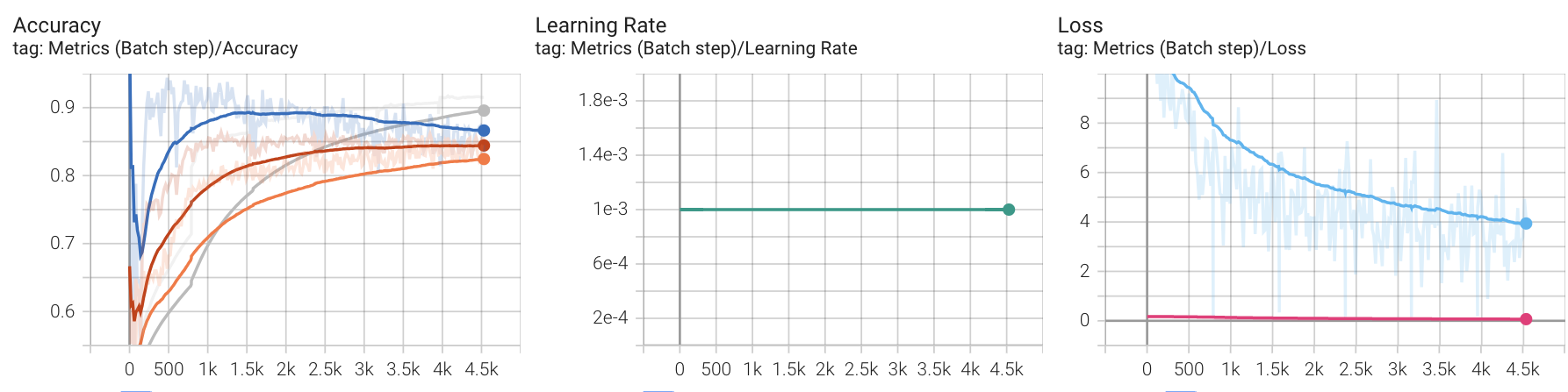}
        \caption{\poolc{} data-set.}
        \label{fig:codegraph_poolc}
    \end{subfigure}
    \hfill
    \begin{subfigure}[b]{\columnwidth}
        \centering
        \includegraphics[width=\columnwidth]{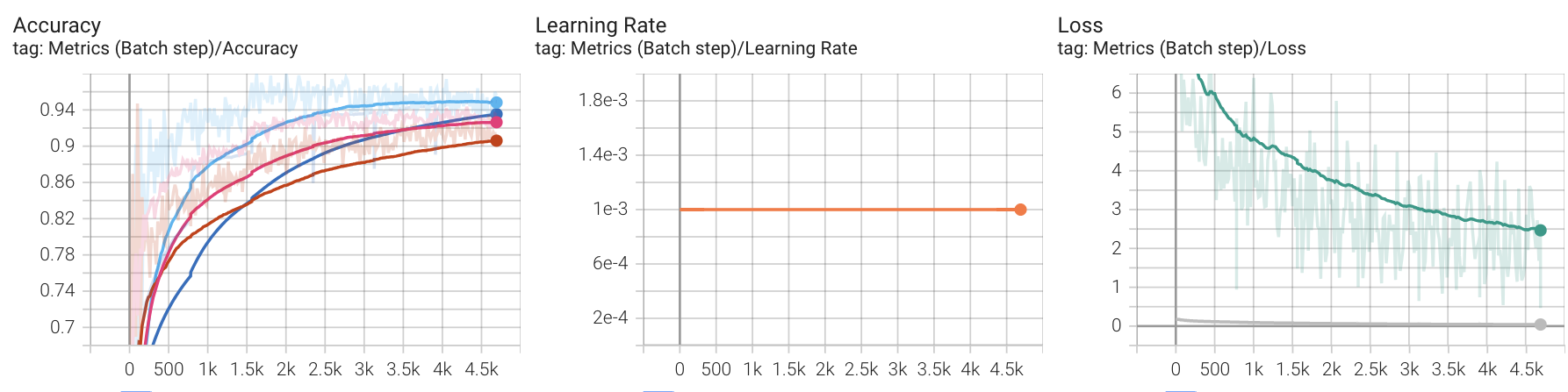}
        \caption{\mixone{} data-set.}
        \label{fig:codegraph_mix_1}
    \end{subfigure}
    \caption{Training curves for \codegraph{} model}
    \label{fig:training_curves_codegraph}
\end{figure}

\newpage
\section{Result Analysis}\label{appendix:results}

\subsection{Confusion Matrix}\label{appendix:results:CM}

\newcommand \cmwidth {0.4}

\subsubsection{BCB Dataset} 
\begin{figure}[H]
        
    \begin{subfigure}[b]{\cmwidth\columnwidth}
        \includegraphics[width=\columnwidth]{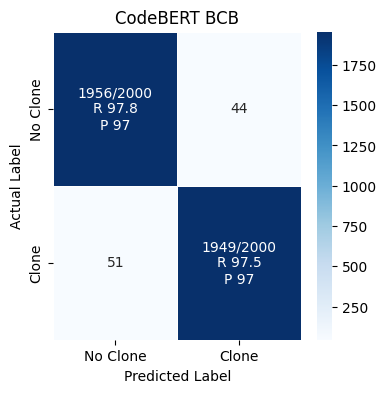}
        \caption{CodeBERT}
        \label{fig:cm_codebert}
    \end{subfigure}
    \hfill
    \begin{subfigure}[b]{\cmwidth\columnwidth}
        \includegraphics[width=\columnwidth]{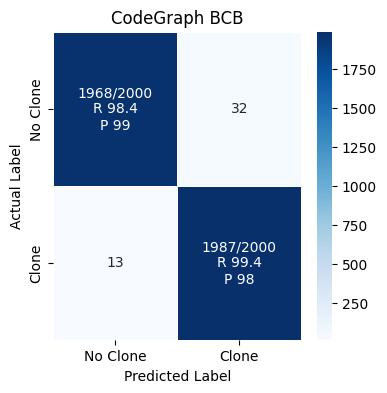}
        \caption{CodeGraph}
        \label{fig:cm_codegraph}
    \end{subfigure}
    \caption{Confusion Matrix : BCB}
    \label{fig:CM_BCB}

\end{figure}

Given in Figure \ref{fig:CM_BCB} are the confusion matrix for CodeBERT and CodeGraph models.

\subsubsection{PoolC Dataset}
\begin{figure}[H]
    \centering
    \begin{subfigure}[b]{\cmwidth\columnwidth}
        \centering
        \includegraphics[width=\columnwidth]{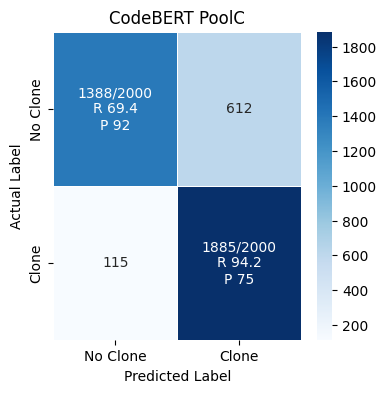}
        \caption{CodeBERT}
        \label{fig:cm_codebert}
    \end{subfigure}
    \hfill
    \begin{subfigure}[b]{\cmwidth\columnwidth}
        \centering
        \includegraphics[width=\columnwidth]{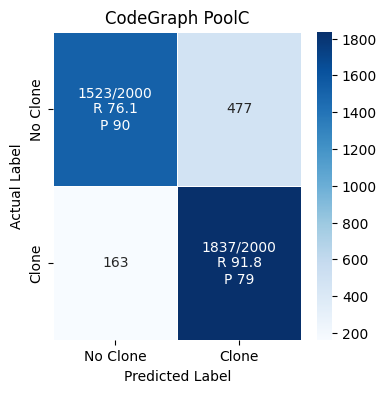}
        \caption{CodeGraph}
        \label{fig:cm_codegraph}
    \end{subfigure}
    \caption{Confusion Matrix : PoolC}
    \label{fig:CM_PoolC}
\end{figure}

Given in Figure \ref{fig:CM_PoolC} are the confusion matrix for CodeBERT and CodeGraph models.

\newpage
\subsection{False Positive Analysis}\label{appendix:results:FPanalysis}

\subsubsection{BCB Dataset} 
\begin{itemize}
    \item There are 18 False positive from both the models combined. Here we observe that the prediction confidence from CodeBERT is very high above 0.9, where as CodeGraph has prediction confidence on a lower side at 0.5 to 0.6. As observed from Code Pair Listings [\ref{code:1.1} \& \ref{code:1.2}], [\ref{code:2.1} \& \ref{code:2.2}]. This suggest that adjusting the classification threshold for CodeGraph can help reduce the False positives which are common in both.

    \item The False Positives from CodeGraph, but True Negative from CodeBERT is seen to be consistently having less confidence, which is below 0.7. Although these False positives are only predicted by CodeGraph, and CodeBERT very strongly predicts them as True Negative. Examples of Code Pair Listing [\ref{code:bcb:1300030} \& \ref{code:bcb:20955454}] following this trend.

    \item The False Positives from CodeBERT, but True Negative from CodeGraph, is seen to have a consistent prediction with confidence less than 0.9. This can be misleading as this is a higher confidence from CodeBERT. on the same side, CodeGraph doesn't have very strong prediction either, but it is atleast consistently predicting them as TN. Example of Code Pair Listing [\ref{code:bcb:5510183} \& \ref{code:bcb:9356670}]
\end{itemize}

\vspace{30px}

\begin{multicols}{2}[\textbf{Code Pair Example} | true\_label : \textbf{NoClone} | pred\_CodeBERT : \textbf{Clone(0.91)} | pred\_CodeGraph : \textbf{Clone(0.62)}]
\begin{lstlisting}[language=Java, caption=code 1, linewidth=.99\columnwidth, label=code:1.1]
public PhoneDurationsImpl(URL url) throws IOException {
    BufferedReader reader;
    String line;
    phoneDurations = new HashMap();
    reader = new BufferedReader(new InputStreamReader(url.openStream()));
    line = reader.readLine();
    while (line != null) {
        if (!line.startsWith("***")) {
            parseAndAdd(line);
        }
        line = reader.readLine();
    }
    reader.close();
}\end{lstlisting}
\begin{lstlisting}[language=Java, caption=code 2, linewidth=.99\columnwidth, label=code:1.2]
public static String getMyGlobalIP() {
    try {
        URL url = new URL(IPSERVER);
        HttpURLConnection con = (HttpURLConnection) url.openConnection();
        BufferedReader in = new BufferedReader(new InputStreamReader(con.getInputStream()));
        String ip = in.readLine();
        in.close();
        con.disconnect();
        return ip;
    } catch (Exception e) {
        return null;
    }
}\end{lstlisting}
\end{multicols}

\vspace{12px}
\begin{multicols}{2}[\textbf{Code Pair Example} | true\_label : \textbf{NoClone} | pred\_CodeBERT : \textbf{Clone(0.91)} | pred\_CodeGraph : \textbf{Clone(0.59)}]
\begin{lstlisting}[language=Java, caption=code 1, linewidth=.99\columnwidth, label=code:2.1]
public static LinkedList<String> read(URL url) throws IOException {
    LinkedList<String> data = new LinkedList<String>();
    HttpURLConnection con = (HttpURLConnection) url.openConnection();
    BufferedReader br = new BufferedReader(new InputStreamReader(con.getInputStream()));
    String input = "";
    while (true) {
        input = br.readLine();
        if (input == null) break;
        data.add(input);
    }
    br.close();
    return data;
}\end{lstlisting}
\begin{lstlisting}[language=Java, caption=code 2, linewidth=.99\columnwidth, label=code:2.2]

protected Reader getText() throws IOException {
    BufferedReader br = new BufferedReader(new InputStreamReader(url.openStream()));
    String readLine;
    do {
        readLine = br.readLine();
    } while (readLine != null && readLine.indexOf("</table><br clear=all>") < 0);
    return br;
}

\end{lstlisting}
\end{multicols}

\vspace{12px}
\begin{multicols}{2}[\textbf{Code Pair Example} | true\_label : \textbf{NoClone} | pred\_CodeBERT : \textbf{NoClone(0.99)} | pred\_CodeGraph : \textbf{Clone(0.65)}]
\begin{lstlisting}[language=Java, caption=code 1, linewidth=.99\columnwidth, label=code:bcb:1300030]

public PhoneDurationsImpl(URL url) throws IOException {

    BufferedReader reader;
    String line;
    phoneDurations = new HashMap();
    
    reader = new BufferedReader(new InputStreamReader(url.openStream()));
    line = reader.readLine();
    
    while (line != null) {
        if (!line.startsWith("***")) {
            parseAndAdd(line);
        }
        
        line = reader.readLine();
    }
    
    reader.close();
}

\end{lstlisting}
\begin{lstlisting}[language=Java, caption=code 2, linewidth=.99\columnwidth, label=code:bcb:20955454]
public void alterarQuestaoMultiplaEscolha(QuestaoMultiplaEscolha q) throws SQLException {
    PreparedStatement stmt = null;
    String sql = "UPDATE multipla_escolha SET texto=?, gabarito=? WHERE id_questao=?";
    try {
        for (Alternativa alternativa : q.getAlternativa()) {
            stmt = conexao.prepareStatement(sql);
            stmt.setString(1, alternativa.getTexto());
            stmt.setBoolean(2, alternativa.getGabarito());
            stmt.setInt(3, q.getIdQuestao());
            stmt.executeUpdate();
            conexao.commit();
        }
    } catch (SQLException e) {
        conexao.rollback();
        throw e;
    }
}
\end{lstlisting}
\end{multicols}

\vspace{12px}
\begin{multicols}{2}[\textbf{Code Pair Example} | true\_label : \textbf{NoClone} | pred\_CodeBERT : \textbf{Clone(0.90)} | pred\_CodeGraph : \textbf{NoClone(0.67)}]
\begin{lstlisting}[language=Java, caption=code 1, linewidth=.99\columnwidth, label=code:bcb:5510183]
public static String getMyGlobalIP() {
    try {
        URL url = new URL(IPSERVER);
        HttpURLConnection con = (HttpURLConnection) url.openConnection();
        BufferedReader in = new BufferedReader(new InputStreamReader(con.getInputStream()));
        String ip = in.readLine();
        in.close();
        con.disconnect();
        return ip;
    } catch (Exception e) {
        return null;
    }
}
\end{lstlisting}
\begin{lstlisting}[language=Java, caption=code 2, linewidth=.99\columnwidth, label=code:bcb:9356670]
private FTPClient loginToSharedWorkspace() throws SocketException, IOException {
    FTPClient ftp = new FTPClient();
    ftp.connect(mSwarm.getHost(), mSharedWorkspacePort);
    if (!ftp.login(SHARED_WORKSPACE_LOGIN_NAME, mWorkspacePassword)) {
        throw new IOException("Unable to login to shared workspace.");
    }
    ftp.setFileType(FTPClient.BINARY_FILE_TYPE);
    return ftp;
}
\end{lstlisting}
\end{multicols}

\newpage
\subsubsection{PoolC Dataset} 
\begin{itemize}
    \item There are 344 total False positives predicted by both the models combine. But here we observe that the confidence is following similar trend with BCB dataset, CodeBERT is having higher prediction confidence of Fasle positive, where as CodeGraph has a lower prediction confidence, with max being at 0.71. Here we see that the positive prediction is majorly coming from confusion of syntactically same keywords present in both, for example having extensive usage of if and for loops. This can be seen in the example Code Pair Listing [\ref{code:poolc:5931} \& \ref{code:poolc:6679}].
    \item The False positive from CodeGraph, but True Negative from CodeBERT is seen to have consistently lower prediction score from CodeGraph, max being 0.78 and average being 0.58. This again suggests that the CodeGraph is understanding the semantics, and with a high classification threshold, should improve significantly. Here the rational behind why the Graph model seem to get them wrong, is it seems to confused on the syntactic structure of the codes. They might not have same keywords, but the structure syntactically is dominating. A code pair Listing [\ref{code:poolc:4851} \& \ref{code:poolc:7175}].
    \item The false positives from CodeBERT, but True negatives from CodeGraph, seems to have stronger prediction of True negatives from Graph models, again showing the Graph models are superior to learn the structural and symatic information with an average score of 0.81. Looking at what might be going wrong with Sequence model would mostly be the keywords having similar names in sequence, but not syntactically similar, as observed in code pair Listing [\ref{code:poolc:2388} \& \ref{code:poolc:1287}].
\end{itemize}

\vspace{12px}
\begin{multicols}{2}[\textbf{Code Pair Example} | true\_label : \textbf{NoClone} | pred\_CodeBERT : \textbf{Clone(0.64)} | pred\_CodeGraph : \textbf{Clone(0.63)}]
\begin{lstlisting}[language=Python, caption=code 1, linewidth=.99\columnwidth, label=code:poolc:5931]
n=int(input())
x=list(map(int,input().split()))
m=10**15
for i in range(101):
    t=x[:]
    s=sum(list(map(lambda x:(x-i)**2,t)))
    m=min(m,s)
print(m)
\end{lstlisting}
\begin{lstlisting}[language=Python, caption=code 2, linewidth=.99\columnwidth, label=code:poolc:6679]
a,b,c,d = map(int, input().split())

ans = -10**18+1
for i in [a,b]:
  for j in [c,d]:
    if ans < i*j: ans = i*j
print(ans)
\end{lstlisting}
\end{multicols}

\vspace{12px}
\begin{multicols}{2}[\textbf{Code Pair Example} | true\_label : \textbf{NoClone} | pred\_CodeBERT : \textbf{NoClone(0.96)} | pred\_CodeGraph : \textbf{Clone(0.60)}]
\begin{lstlisting}[language=Python, caption=code 1, linewidth=.99\columnwidth, label=code:poolc:4851]
import sys
x=int(input())

n=1
while(100*n<=x):
    if(x<=105*n):
        print(1)
        sys.exit()
    n+=1
print(0)
\end{lstlisting}
\begin{lstlisting}[language=Python, caption=code 2, linewidth=.99\columnwidth, label=code:poolc:7175]
s = str(input())
t = str(input())
revise = 0
len_str = len(s)
for i in range(len_str):
    if s[i] != t[i]:
        revise += 1
print(int(revise))
\end{lstlisting}
\end{multicols}

\vspace{12px}
\begin{multicols}{2}[\textbf{Code Pair Example} | true\_label : \textbf{NoClone} | pred\_CodeBERT : \textbf{Clone(0.83)} | pred\_CodeGraph : \textbf{NoClone(0.93)}]
\begin{lstlisting}[language=Python, caption=code 1, linewidth=.99\columnwidth, label=code:poolc:2388]
K, N= map(int, input().split())
A = list(map(int, input().split()))
max=K-(A[N-1]-A[0])
for i in range(N-1):
    a=A[i+1]-A[i]
    if max<a:
        max=a
print(K-max)
\end{lstlisting}
\begin{lstlisting}[language=Python, caption=code 2, linewidth=.99\columnwidth, label=code:poolc:1287]
x = float(input())
if 1 >= x >= 0:
    if x == 1:
        print(0)
    elif x == 0:
        print(1)
\end{lstlisting}
\end{multicols}

\newpage
\subsection{False Negative Analysis}\label{appendix:results:FNanalysis}

\subsubsection{BCB Dataset} 
\begin{itemize}
    \item There is zero overlap of False Negative between both the models. This is also due to the fact that there are very low false negative overall, due to the model tending to overfit on the dataset.  
    \item The False Negatives from the CodeGraph, but True Positives from CodeBERT, shows consistently lower confidence at an average of 0.7. This shows that we can tweak the classification threshold to handle these lower confidence scores. On further inspection of these cases, which where just 14, shows that this prediction of false negatives are more cause of variation in parameters, which seems to mislead the model to not detect them as clone. Morover we can argue these are Type IV clones, which would be better identified, given more context. An example can be seen in the Code Pair Listing [\ref{code:bcb:23677124} \& \ref{code:bcb:23677129}]
    \item The False Negatives from the CodeBERT, but True Positives from CodeGraph, have a strong very high confidence. This is not helpful, as it is clearly seen that the sequence model is predicting them wrongly as Negataives with a conf average of 0.99. This is a very intersting case, as all the 52 of these cases seems to have the code very different syntactically, but semantically they are same. This shows how the graph model has an edge over the sequence model. This can be seen in the Code Pair Listing [\ref{code:bcb:3257108} \& \ref{code:bcb:21044331}]
\end{itemize}

\vspace{12px}
\begin{multicols}{2}[\textbf{Code Pair Example} | true\_label : \textbf{Clone} | pred\_CodeBERT : \textbf{Clone(0.91)} | pred\_CodeGraph : \textbf{NoClone(0.58)}]
\begin{lstlisting}[language=Java, caption=code 1, linewidth=.99\columnwidth, label=code:bcb:23677124]
public FTPClient sample1c(String server, int port, String username, String password) throws SocketException, IOException {
        FTPClient ftpClient = new FTPClient();
        ftpClient.setDefaultPort(port);
        ftpClient.connect(server);
        ftpClient.login(username, password);
        return ftpClient;
}\end{lstlisting}
\begin{lstlisting}[language=Java, caption=code 2, linewidth=.99\columnwidth, label=code:bcb:23677129]
public FTPClient sample3b(String ftpserver, String proxyserver, int proxyport, String username, String password) throws SocketException, IOException {
        FTPHTTPClient ftpClient = new FTPHTTPClient(proxyserver, proxyport);
        ftpClient.connect(ftpserver);
        ftpClient.login(username, password);
        return ftpClient;
}\end{lstlisting}
\end{multicols}

\vspace{12px}
\begin{multicols}{2}[\textbf{Code Pair Example} | true\_label : \textbf{Clone} | pred\_CodeBERT : \textbf{NoClone(0.99)} | pred\_CodeGraph : \textbf{Clone(0.97)}]
\begin{lstlisting}[language=Java, caption=code 1, linewidth=.99\columnwidth, label=code:bcb:3257108]

public static String getMD5(String s) {

    try {
        MessageDigest m = MessageDigest.getInstance("MD5");
        m.update(s.getBytes(), 0, s.length());
        s = new BigInteger(1, m.digest()).toString(16);
    } 
    catch (NoSuchAlgorithmException ex) {
        ex.printStackTrace();
    }
    return s;
}

\end{lstlisting}
\begin{lstlisting}[language=Java, caption=code 2, linewidth=.99\columnwidth, label=code:bcb:21044331]
private static byte[] getKey(String password) throws UnsupportedEncodingException, NoSuchAlgorithmException {
    MessageDigest messageDigest = MessageDigest.getInstance(Constants.HASH_FUNCTION);
    messageDigest.update(password.getBytes(Constants.ENCODING));
    byte[] hashValue = messageDigest.digest();
    int keyLengthInbytes = Constants.ENCRYPTION_KEY_LENGTH / 8;
    byte[] result = new byte[keyLengthInbytes];
    System.arraycopy(hashValue, 0, result, 0, keyLengthInbytes);
    return result;
}\end{lstlisting}
\end{multicols}

\newpage
\subsubsection{PoolC Dataset} 
\begin{itemize}
    \item There are 34 False Negatives predicted by both the models combined. Here we see that the average score from Graph model is 0.64 where as 0.87 is the average score from the sequence model. This shows how the sequence model is very confidently wrong, which is not a good sign although, when examples are examined for this case, we find the clone pairs distinctly have a difference in length in the code, which seems to be the reason for these wrong predictions. Code Pair Listing [\ref{code:poolc:7785} \& \ref{code:poolc:4401}] demonstrates this. 
    \item The False Negatives from CodeGraph, but True positives from CodeBERT, shows a consistently lower score of confidence with average confidence being 0.63. where as the true positives as well from codeBERT have lower confidence average of 0.78. Here the CodeGraph is marignaly doing wrong, and this seems to be due to the code length again. the difference in the size of code snippet is larger. This can be seen again from the Code Pair Listing  [\ref{code:poolc:5696} \& \ref{code:poolc:3165}].
    \item The False Negatives from CodeBERT, but True Positives from CodeGraph, are around 83. This cases seems to be strongly False at an average score of 0.81 for the CodeBERT, which is not a good sign of the model predicting them wrong confidently, again, the reason looks the same as the snippet sizes being very different. Simialarly average confidence of True positive from CodeGraph is 0.61, which is again not that confident. This can be seen with the Code Pair Listing  [\ref{code:poolc:6612} \& \ref{code:poolc:8682}].
\end{itemize}

\vspace{12px}
\begin{multicols}{2}[\textbf{Code Pair Example} | true\_label : \textbf{Clone} | pred\_CodeBERT : \textbf{NoClone(0.88)} | pred\_CodeGraph : \textbf{NoClone(0.96)}]
\begin{lstlisting}[language=Python, caption=code 1, linewidth=.99\columnwidth, label=code:poolc:7785]


while True: 

    a = input()
    
    if a == '0':
        break
    
    print(sum(map(int,*a.split())))
\end{lstlisting}
\begin{lstlisting}[language=Python, caption=code 2, linewidth=.99\columnwidth, label=code:poolc:4401]
n = int(input())
res = 0
while n != 0:
 res = n%10
 dropped = n
 while dropped//10 != 0:
  dropped = dropped//10
  res += dropped%10
 print(res)
 res = 0
 n = int(input())
\end{lstlisting}
\end{multicols}

\vspace{12px}
\begin{multicols}{2}[\textbf{Code Pair Example} | true\_label : \textbf{Clone} | pred\_CodeBERT : \textbf{Clone(0.52)} | pred\_CodeGraph : \textbf{NoClone(0.59)}]
\begin{lstlisting}[language=Python, caption=code 1, linewidth=.99\columnwidth, label=code:poolc:5696]
while True:
    
    n = input()
    
    if n == "0":
        break
        
    print(sum([int(i) for i in n]))
\end{lstlisting}
\begin{lstlisting}[language=Python, caption=code 2, linewidth=.99\columnwidth, label=code:poolc:3165]
while True:
    num = input()
    if int(num) == 0:
        break
    sum = 0
    for i in num:
        a = int(i)
        sum += a
    print(sum)\end{lstlisting}
\end{multicols}

\vspace{12px}
\begin{multicols}{2}[\textbf{Code Pair Example} | true\_label : \textbf{Clone} | pred\_CodeBERT : \textbf{NoClone(0.62)} | pred\_CodeGraph : \textbf{Clone(0.66)}]
\begin{lstlisting}[language=Python, caption=code 1, linewidth=.99\columnwidth, label=code:poolc:6612]
H,A = map(int,input().split())
cnt = 0
while True:
    if H <= 0:
        print(cnt)
        break
    else:
        H -= A
        cnt += 1\end{lstlisting}
\begin{lstlisting}[language=Python, caption=code 2, linewidth=.99\columnwidth, label=code:poolc:8682]
h,a = map(int, input().split())
an, bn = divmod(h,a)
if bn == 0:
    print(an)
else:
    print(an+1)
\end{lstlisting}
\end{multicols}




\end{document}